\documentclass[conference]{IEEEtran}
\ifCLASSINFOpdf
\else
\fi
\usepackage[utf8]{inputenc}

\usepackage{times}
\usepackage{comment}
\usepackage{amssymb}
\usepackage{algorithm}
\usepackage[noend]{algpseudocode}
\usepackage{setspace}
\usepackage{xspace}
\usepackage{graphicx}
\usepackage{subcaption}
\usepackage{amsmath}
\usepackage{color}

\usepackage{multirow}
\usepackage{tabularx,booktabs}
\usepackage{verbatimbox}
\usepackage[all=normal, title=normal, paragraphs=tight, wordspacing=normal]{savetrees}

\newcommand{\soa}{stochastic robust approximation\xspace}
\newcommand{\Soa}{Stochastic robust approximation\xspace}
\newcommand{\SOA}{Stochastic Robust Approximation\xspace}

\newcommand{\dmt}{dynamic mixed training\xspace}

\newcommand{\DMT}{Dynamic Mixed Training\xspace}

\newcommand{\soov}{sound over-approximations of violations\xspace}

\newcommand{\sys}{MixTrain\xspace}

\DeclareMathOperator*{\argmax}{arg\,max}

\DeclareMathOperator*{\E}{\mathbb{E}}

\renewcommand{\baselinestretch}{0.99}

\begin{document}

\title{MixTrain: Scalable Training of Verifiably Robust\\ Neural Networks \vspace{-10pt}}

\author{\IEEEauthorblockN{Shiqi Wang\IEEEauthorrefmark{1},
		Yizheng Chen\IEEEauthorrefmark{1}, Ahmed Abdou\IEEEauthorrefmark{2} and
		Suman Jana\IEEEauthorrefmark{1}}
	\IEEEauthorblockA{\IEEEauthorrefmark{1}Columbia University}
	\IEEEauthorblockA{\IEEEauthorrefmark{2}Pennsylvania State University\vspace{-10pt}}
	}

\maketitle

\begin{abstract}
\label{sec:abst}
Making neural networks robust against adversarial inputs has resulted in an arms race between new defenses and attacks that break them. The most promising defenses, adversarially robust training and verifiably robust training, have limitations that severely restrict their practical applications. The adversarially robust training only makes the networks robust against a subclass of attackers (e.g., first-order gradient-based attacks), leaving it vulnerable to other attacks. In this paper, we reveal such weaknesses of adversarially robust networks by developing a new attack based on interval gradients. By contrast, verifiably robust training provides protection against any L-p norm-bounded attacker but incurs orders of magnitude more computational and memory overhead than adversarially robust training. Therefore, the usage of verifiably robust training has so far been restricted only to small networks.

In this paper, we propose two novel techniques, \soa and \dmt, to drastically improve the efficiency of verifiably robust training without sacrificing verified robustness. Our techniques leverage two critical insights: (1) instead of computing expensive sound over-approximations of robustness violations over the entire training set, sound over-approximations over randomly subsampled training data points are sufficient for efficiently guiding the robust training process. Sound over-approximation over the entire training dataset are only required for measuring the verified robustness of the final network; and  (2) We observe that the test accuracy and verifiable robustness often conflict (i.e., increasing one decreases the other)   during the training process after certain training epochs. Therefore, we use a dynamic loss function to adaptively balance the accuracy and verifiable robustness of a network for each epoch.

We designed and implemented our techniques as part of \sys and thoroughly evaluated it on six different networks trained on three popular datasets including MNIST, CIFAR, and ImageNet-200. Our detailed evaluations show that \sys can achieve up to 95.2\% verified robust accuracy against $L_\infty$ norm-bounded attackers while taking $15$ and $3$ times less training time than state-of-the-art verifiably robust training and adversarially robust training schemes, respectively. Furthermore, \sys easily scales to larger networks like the one trained on ImageNet-200 dataset, significantly outperforming the existing verifiably robust training methods.

\end{abstract}

\IEEEpeerreviewmaketitle

\section{Introduction}
\label{sec:intro}

Neural networks have been shown to be vulnerable against carefully crafted adversarial inputs~\cite{szegedy2013intriguing}. Such vulnerabilities can have potentially fatal consequences especially in security-critical systems such as autonomous vehicles~\cite{teslacrash1,teslacrash2,teststruggle} and unmanned aircrafts~\cite{julian2016policy,MQ-4C,reluval2018,katz2017reluplex}. Unfortunately, most of the proposed defenses provide no clear security guarantee and have been consistently broken by increasingly stronger attacks resulting in an arms race~\cite{papernot2015distillation,cisse2017parseval,pixeldp, papernot2017practical,moosavi2016deepfool,carlini2017towards}. 

So far, only two principled robust training methods (adversarially robust training and verifiably robust training) have been found to be robust against a wide range of attacks. However, both of them suffer from significant limitations severely restricting their practical applicability. The adversarially robust training only makes the networks robust against a specific subclass of attackers (e.g., known attacks using first-order gradients), leaving it potentially vulnerable to other attacks. This is not just a hypothetical concern. In this paper, we demonstrate a new class of first-order interval gradient attacks that cause up to 40\% increase in attack success rate against state-of-the-art adversarially robust networks. By contrast, verifiably robust training provides sound protections against any L-p norm-bounded attackers but incurs orders of magnitude more computational and memory overhead than adversarially robust training. Therefore, the usage of verifiably robust training so far has mostly been restricted to only small networks. 

In this paper, we solve these challenges by designing \sys, a scalable verifiably robust training technique that can provide strong verifiable robustness against any L-p norm-bounded attacker with significantly faster training time than any existing robust (both adversarial and verifiable) training schemes. Essentially, \sys gives the best of both adversarially and verifiably robust training, i.e., protection against a strong attacker while still keeping the training process efficient. Below, we first summarize the main causes behind the scalability of existing verifiable robust training schemes and then describe how we address these challenges.

All robust training schemes essentially add a robustness component to the loss function that provides an estimate of robustness violations an attacker can potentially induce for the current state of the network. The robust loss (in some cases together with regular loss) is then minimized by searching for the appropriate network weights during the robust training process. The main distinguishing factor between adversarially robust and verifiably robust training is the method used to compute the robust loss during the training process. Adversarially robust training uses different existing attacks to get a lower bound on the robustness violations and therefore can only defend against a specific (or a specific class of) attack(s) even when the training process achieves low robust loss value. By contrast, verifiably robust training uses sound over-approximation methods like symbolic interval analysis~\cite{reluval2018,shiqi2018efficient}, abstract interpretation~\cite{gehrai,mirman2018differentiable}, and robust optimization techniques~\cite{wong2018provable} to find a sound upper bound on the robustness violations that can be triggered by an L-p norm-bounded attacker. 

We observe that existing verifiably robust training techniques suffer from two major limitations. First, during each training epoch (i.e., one pass over the training data), the sound over-approximation methods are run for each training data point to estimate the robustness violations given the current set of weights of the network. This is an extremely computation- and memory-intensive process. Second, the robust loss and regular loss often conflict during the robust training process, i.e., decreasing one increases the other (as shown in Section~\ref{subset:problem}). Such conflicts prevent verifiably robust training from efficiently achieving high test accuracy.

We propose two novel techniques, \soa and \dmt, to address the issues mentioned above to make verifiably robust training significantly more efficient. First, we observe that sound over-approximation of robustness violations for each training data point is not necessary during training to learn a network with high verifiable robustness. Sound over-approximations of robustness violations over a randomly selected subset of the training data can still make the robust training process converge to a verifiably robust network over all the training and test data. This approach drastically cuts down the training overhead. Second, we use a dynamic loss function to adaptively balance the robust loss and regular loss for each epoch that can increase both test accuracy and verifiable robustness simultaneously. 

We implemented and thoroughly evaluated \sys on six different neural networks trained on three popular datasets including MNIST, CIFAR, and ImageNet-200. While our techniques are generic and can integrate with any sound over-approximation method, in this paper, we used symbolic intervals~\cite{shiqi2018efficient} to soundly over-approximate the number of robustness violations. Our detailed evaluations show that \sys can achieve up to 95.2\% verified robust accuracy against $L_\infty$ norm-bounded attackers while taking $15\times$ and $3\times$ less training time than state-of-the-art verifiably robust training~\cite{wong2018scaling} and adversarially robust training schemes~\cite{madry2017towards}, respectively. Furthermore, \sys easily scales to larger networks trained on ImageNet-200 dataset and significantly outperforms earlier robust training methods. \sys also requires $10\times$ less memory, on average, than existing verifiably robust training schemes.

Our key contributions are summarized below:

\begin{itemize}
\item We present a new first-order interval gradient attack that causes up to 40\% more violations in the state-of-the-art adversarially robust networks.

\item We propose two novel techniques (i.e., \soa and \dmt) for making the verifiably robust training process highly efficient and effective at minimizing both robust loss and regular loss simultaneously. 

\item We implemented our techniques as part of \sys and thoroughly evaluated on six different networks trained on three popular datasets including MNIST, CIFAR, and ImageNet-200. 
\sys can achieve up to 95.2\% verified robust accuracy against $L_\infty$ norm-bounded attackers while taking $15\times$ and $3\times$ less training time than state-of-the-art verifiably and adversarially robust methods, respectively.
\end{itemize}

\section{Background}
\label{sec:background}
In this section, we provide an overview of how robust training of Neural Networks (NNs) can be formulated as a robust optimization problem to clearly explain the challenges behind robust training. We further describe three different robustness definitions (adversarial~\cite{madry2017towards,sinha2018certifying}, distributional~\cite{sinha2018certifying}, and verifiable robustness~\cite{wong2018scaling,mirman2018differentiable,dvijotham2018training}) that assume different threat models and summarize different existing training schemes for making NNs robust under these definitions and their limitations. 

\subsection{Robust Optimization}
\label{subsec:robust_optimization}

Robust optimization techniques broadly try to find solutions that are robust under certain bounded perturbations in different parameters of the target optimization problem. Training an NN that is robust to adversarial inputs, created by adding some bounded perturbations to benign inputs, can also be represented as a robust optimization problem. Formally, a NN $f_{\theta}:\mathbb{R}^d\rightarrow \mathbb{R}^k$ maps
$d$ input features to $k$ output features, with weights $\theta$.
Given the input pair $(x,y)$ drawn from the underlying distribution $\mathcal{D}$,
$f_{\theta}$ predicts the label of $x$ as $\hat{y} = \argmax_{\{\hat{y}_{1},...,\hat{y}_{k}\}}{f_\theta(x)}$. 
The correctness, robustness, violation, and robust accuracy of $f_{\theta}$ on $(x,y)$ are defined as follows.
\begin{itemize}
	\item {\bf Correctness:} $x$ is classified \emph{correctly} if $\hat{y} = y$, where $y$ is the true label.
	\item {\bf Robustness:} $f_\theta$ is \emph{robust} in the robustness region $B_\epsilon(x)$ around $x$ if $\forall \tilde{x}\in B_\epsilon(x)$ 
	$f_\theta$ always computes $\argmax_{\hat{y}} f_\theta(\tilde{x})=y$.
	\item {\bf Violation:} If $\exists \tilde{x}\in B_\epsilon(x)$, such that $\argmax_{\hat{y}} f_\theta(\tilde{x}) \neq y$,
	$\tilde{x}$ is a \emph{violation} of the robustness property.
	\item {\bf Robust accuracy:} The expected percentage of test samples over the underlying distribution D that is both correctly classified and have no violations within the corresponding robustness region $B_\epsilon$.
\end{itemize}

The most common way to define a robustness region ($B_\epsilon(x)$) is to use a $L_\infty$-ball of radius $\epsilon$ around $x$~\cite{szegedy2013intriguing}. However, the definitions presented above also apply to any arbitrary $B_\epsilon$ including other $L_p$ norm-bounded balls.



The regular NN training process involves the objective function as shown in Eqn. \ref{eq:re}, which optimizes the weights $\theta$ to minimize the expected risk of the loss value $\E_{(x,y)\sim\mathcal{D}}(L(\theta, x, y))$. The loss function $L(\theta, x, y)$ is used to measure the performance of the network on a specific input $(x, y)$. A correct prediction results in a very small loss value while incorrect predictions cause the loss values to be large. The cross-entropy loss function is commonly used to train the network. Since the underlying input distribution $\mathcal{D}$ is unknown, the standard Empirical Risk Minimization (ERM)~\cite{vapnik1999overview,vapnik1992principles} method uses the empirical distribution $\mathcal{D}_0$, as represented by the training dataset of size $N$, to minimize the loss value $ \E_{(x,y)\sim\mathcal{D}_0}(L(\theta, x, y)) = \frac{1}{N}\sum{L(\theta, x, y)}$.

\vspace{-10pt}
\begin{equation}
\text{Regular Training Obj.}\qquad\quad\quad\min_\theta \E_{(x,y)\sim\mathcal{D}}(L(\theta, x, y))
\label{eq:re}
\end{equation}
\begin{equation}
\text{Robust Training Obj.}\quad\min_\theta \max_{\tilde{x}\in B_\epsilon(x)} \E_{(x,y)\sim\mathcal{D}}(L(\theta, \tilde{x}, y))
\label{eq:ro}
\end{equation}

By contrast, the robust training of NNs~\cite{huang2015learning, shaham2015understanding, madry2017towards}  minimizes the empirical risk of the largest loss value from $B_\epsilon$ (i.e., maximizes robust accuracy). Eqn. \ref{eq:re} and Eqn. \ref{eq:ro} show the differences between the objective functions of regular and robust training. Specifically, The NN robust training tries to solve two related problems as shown by the objective function in Eqn.~
\ref{eq:ro}.

\begin{itemize}
	\item {\bf Inner Maximization Problem}: Finding $\tilde{x}$ to maximize the loss value within a robustness region  $B_\epsilon(x)$.
	\item {\bf Outer Minimization Problem}: Optimizing weights $\theta$ to minimize the empirical maximal loss.
\end{itemize}
The inner maximization problem of the robust training procedure is known to be NP-hard~\cite{katz2017reluplex}. Existing robust training approaches use either gradient-based attacks or sound over-approximations to find approximate solutions. The gradient-based attacks used by adversarially robust training schemes provide an under-approximation of the optimal solution,  while verifiably robust training uses sound over-approximation techniques. Both of these training schemes use standard gradient descent and backpropagation for solving the outer minimization problem. We describe the details of each type of robustness, the corresponding threat models, and training procedures in detail in the rest of this section. 

Note that just like the regular training process, robust training process also uses the empirical distribution $\mathcal{D}_0$ represented by the training dataset to compute the solutions to the inner/outer problems and expect that the solution will generalize to the underlying distribution. Also, both regular and robust training, instead of updating the NN weights for each training data sample separately, in practice, use batch training to make the training process more computationally efficient. The training data is split into batches of fixed size and for each batch of training data points $X=\{x_1,...,x_m\}$ and labels $Y=\{y_1, ...,y_m\}$, the batch training procedure computes the loss value and performs back-propagation to update the weights to minimize the loss value.

\subsection{Adversarial Robustness}
\label{subsec:rg}

\noindent\textbf{Threat model.}
Besides bounding the perturbations by an attacker within the robustness region $B_\epsilon(x)$, adversarial robustness further assumes that the attacker is restricted to a specific subclass of attacks, i.e., first-order gradient-guided attackers. Recent works have trained NNs to be robust against different existing first-order attacks like Projected Gradient Descent (PGD) and Carlini-Wagner (CW)~\cite{kurakin2016adversarial,goodfellow2014explaining,tramer2017ensemble,ducoffe2018adversarial}. 
 Unfortunately, all of these defenses have been shown to be easily broken by slightly changing the corresponding attack exposing the inherent limitations of the underlying robust training schemes~\cite{athalye2018obfuscated,lucas2018towards}.  

To avoid such issues, Madry et al.~\cite{madry2017towards} recently proposed a defense scheme against all first-order adversaries  where the attacker is assumed to use first order gradients of the NN in arbitrary ways but not use any higher-order gradients. Madry et al. justified the choice of their threat model by arguing that computational complexity of higher-order gradients is not clearly understood and thus the memory and computation cost may be significantly higher than computing first-order gradients. We summarize their robust training scheme below.

\noindent\textbf{Adversarially robust training.}
The state-of-the-art adversarially robust training scheme of Madry et al.~\cite{madry2017towards} relies on
first-order gradient-guided PGD attacks to search for the largest loss value within $B_\epsilon$. In fact, given input pair ($x$, $y$), first-order attacks to estimate the lower bound of the inner maximization problem:
\begin{equation}
L(Attack(x), y)\leq \max_{\tilde{x}\in B_\epsilon(x)} L(f_\theta(\tilde{x}), y)
\end{equation}

Specifically, for a batch of training data ($X$, $Y$), the training process first conducts PGD perturbations with a random starting point to generate a perturbed training batch $X'=PGD(X)$, representing the estimated solution to the inner problem $\max_{\tilde{x}\in B_\epsilon(x)} L(f_\theta(\tilde{x}), y)$. The corresponding losses of $X'$ and labels $Y$ are then used to solve the outer minimization problem (Eqn. \ref{eq:ro}) using gradient descent by minimizing the function shown below.
\begin{equation}
\min_\theta \E_{(x,y)\sim\mathcal{D}}L(f_\theta(PGD(x)),y)
\end{equation}

Madry et al.'s scheme is relatively efficient (though more expensive than regular training) and has been shown to be robust against known first-order attacks like Projected Gradient Descent (PGD)~\cite{kurakin2016adversarial}, Fast Gradient Sign Method (FGSM)~\cite{goodfellow2014explaining}, and Carlini-Wagner (CW) attacks~\cite{carlini2017towards}. In this paper, we define the \emph{Estimated Robust Accuracy (ERA)} of a network as the percentage of test samples that are robust against known first-order attacks like PGD.

\noindent\textbf{Limitations.}
Besides ignoring attacks using second- and higher-order derivatives, Madry et al.'s scheme also assumes that starting a first-order attack like PGD from a random starting point for each batch can provide a good solution to the inner maximization problem. To support this claim, they present experimental results from $100,000$ random starting points showing that all solutions to the inner maximization problem found by PGD in these cases are distinct local optimas, yet they have very similar loss values. Due to the concentrated distribution of the loss values, Madry et al. concluded that the Projected Gradient Descent (PGD)
attacker with a random start can be thought as the ``ultimate'' first-order adversary and thus can provide a good solutions to the inner maximization problem that can be achieved by arbitrary first-order attackers.

However, as NNs are non-convex, the solutions to the inner maximization problem found by a first-order attack like PGD can vary widely based on the starting point. Moreover, the non-convex nature of the NNs often make the {\it good} starting points to be very hard to find using random sampling used in Madry et al.'s experiment described earlier. In Section \ref{sec:eval_ia}), we present a new attack that uses first-order interval gradients to carefully good promising starting points for existing attacks like PGD. We show in Section~\ref{sec:eval_ia} that the estimated \emph{adversarial robust loss} value $L(f_\theta(PGD(x)), y)$ using PGD tend to be much smaller than the true maximal loss, as we will show in Section~\ref{sec:eval_ia}. Moreover, we find that adapting the adversarially robust training process to use  interval attacks converge significantly slowly than the corresponding PGD version. This result exposes the fundamental limitation of such training methods. By contrast, our verifiably robust training scheme are  more efficient and provides significantly stronger 
guarantees against stronger attackers than adversarially robust training. 

\subsection{Distributional Robustness}

\noindent\textbf{Threat model.}
Instead of defining the robustness region $B_\epsilon$ in terms of L-p norms, distributional robustness allows the adversary to draw any data samples (i.e., images) from a distribution $P$ as long as it is within a bounded Wasserstein distance of $\rho$ from the actual underlying data distribution $\mathcal{D}$~\cite{sinha2018certifying}. Formally, the robustness region is $\mathcal{P} = \{P: W_c(P, \mathcal{D}) \leq \rho\}$. There is no other restriction on the information that the adversary has access to or the types of attack the adversary can perform.

\noindent\textbf{Training scheme.}
Unfortunately, solving the inner maximization problem involving the Wasserstein distance is intractable. Therefore, To make the problem tractable, Sinha et al.~\cite{sinha2018certifying} used the Lagrangian relaxation to reformulate the distribution distance bound into a $L_2$ distance penalty. Since $\mathcal{D}$ is unknown, the authors used the empirical distribution $\mathcal{D}_0$ for computing the solutions.

\noindent\textbf{Limitations.}
The distributional robust training provides an useful upper bound on the expectation of the loss function only when the bounded robustness region $B_\epsilon$ is very small (e.g., $||\epsilon||_2\leq 0.1$). Such training become significantly ineffective as the bound becomes larger (e.g., $||\epsilon||_\infty\leq 0.1$). Moreover, the training methods require the NNs to have smooth activation functions like ELU and thus cannot support any networks using ReLUs, one of the most popular activation functions. 

\subsection{Verifiable Robustness}


\noindent\textbf{Threat model.}
Unlike the other two robustness definitions, verifiable robustness can provide protection against
attackers with unlimited computation/information as long as the perturbations are limited within a 
robustness region $B_\epsilon(x)$.
The attacker can use any information, draw input from arbitrary distributions, and launch unknown attacks. 

\noindent\textbf{Verifiably robust training.}
\label{subsec:sa}
To capture the capabilities of unbounded attackers, sound analysis techniques are used to over-estimate
the solutions to the inner maximization problem for a given the robustness region $B_\epsilon$. The soundness ensures that no successful attack can be constructed within its $B_\epsilon$ input range if the analysis found no violations. Existing sound analysis techniques include convex polytope~\cite{wong2018provable}, symbolic intervals~\cite{reluval2018, shiqi2018efficient}, abstract domains~\cite{gehrai}, Lagrangian relaxation~\cite{dvijotham2018dual} and relaxation with Lipschitz constant~\cite{weng2018towards}. Essentially, all of these perform a sound transformation $T_\epsilon$ from the input to the output of the network $f_\theta$. Formally, given input $x \in \mathbb{X}$ and allowable input range $B_\epsilon(x)$, the transformation $T_\epsilon$ is sound if following condition is true:
\begin{equation}
\{f_\theta(\tilde{x})|\tilde{x}\in B_\epsilon(x), \forall x \in \mathbb{X}\}\subseteq T_\epsilon(\mathbb{X})
\end{equation}

The verifiable robustness for an input pair $(x,y)$ is defined as  $\argmax_{j\in\{1,...,k\}}d_\epsilon(x) = y$. It indicates that the output for class $y$ for input $x$ will always be predicted as the largest among all outputs estimated by sound over-approximation $T_\epsilon(x)$. Here, $d_\epsilon(x)$ denotes the worst-case outputs produced by $T_\epsilon(x)$, which corresponds to the estimated worst regular loss value $L(d_\epsilon(x),y)$ within input range $B_\epsilon(x)$. Such sound over-approximations provide the upper bound of the inner maximization problem:

\begin{equation}
\max_{\tilde{x}\in B_\epsilon(x)} L(f_\theta(\tilde{x}), y)\leq L(d_\epsilon(x),y)
\end{equation}


For a batch of training data points ($X$, $Y$), the training process first computes the
\emph{verifiable robust loss} $L(d_\epsilon(X), Y)$ using sound over-approximation methods. Therefore, the outer minimization problem in Eqn. \ref{eq:ro} becomes:
\begin{equation}
\min_\theta \E_{(x,y)\sim\mathcal{D}} L(d_\epsilon(x),y)
\label{eq:rt}
\end{equation}

Such training increases the \emph{Verified Robust Accuracy (VRA)} of a NN, the percentage of test samples within $B_\epsilon$ that can be verified using sound over-approximation to have no violations. The VRA provides a sound lower bound on the robust accuracy of a network as test samples might fail to verify due to false positives generated by over-approximation. 

\noindent\textbf{Limitations.}
While verifiably robust training schemes tend to achieve both high ERA and high VRA, existing training schemes are prohibitively expensive both in terms of computations and required memory, being hundreds of times slower than normal training methods. Moreover, due to the over-approximated violations, existing training schemes also sacrifice test accuracy to improve robustness. Such problems prevent these methods to be applied to real-world applications. In this paper, we specifically focus on verifiably robust training and propose two novel techniques that significantly cuts down the overhead of verifiably robust training while still achieving high accuracy and VRA.

\section{Interval attack}
\label{sec:ia}


In this section, we describe a novel interval-gradient-based attack against the state-of-the-art adversarially robust neural 
networks trained using Madry et al.'s~\cite{madry2017towards} method. Their technique uses PGD attacks to find a maximal loss value for the inner maximization problem shown in Eqn.~\ref{eq:ro}. As PGD is a first-order gradient-guided attack and the neural networks are non-convex, the maximal value found by the PGD might not be the optimal solution (i.e., the maximum) to the inner maximization problem. However, Madry et al. claimed that PGD attacks are the ``ultimate'' first-order attacks, i.e., no other first-order attack will be able to significantly improve over the solutions found by PGD. In order to support their claim, Madry et al. performed $100,000$ iterations of the PGD attack from random starting points within bounded $L_{\infty}$-balls of test inputs and showed that all the solutions found by these instances are distinct local optimas with similar loss values. Therefore, they speculated that these distinct local optimas found by PGD attacks are very close to the best solution that can be found by any first-order attacker.

In this paper, we demonstrate that Madry et al.'s  assumption that the solutions found by the PGD attacks cannot be improved by only using first-order information to be flawed. Our first-order interval-gradient-based attack can effectively find {\it good} starting points for PGD attacks that significantly improves the attack success rate (up to 40\% more) as opposed to starting from random points. Our experiments found that such starting points are not distributed evenly due to the non-convex nature of the neural networks and therefore are very hard to find using random sampling. Our attack demonstrates the pitfalls behind making assumptions about a neural network's behavior, given its non-convexity, based on random sampling. 

We further demonstrate that extending adversarial training method to use the interval attacks struggles to converge well (as shown in Figure~\ref{fig:ia_adv}) even after training for a extremely long time (12 hours) for a very small MNIST networks. We believe that this shows the fundamental limitations of adversarially robust training when used together with advanced attacks. By contrast, our verifiably robust training scheme does not make any such assumptions about the attacker and is more efficient than adversarially robust raining schemes as shown in Section~\ref{sec:eval}.

\noindent{\bf Intuition behind interval attacks.} Due to the non-convexity of the neural networks, existing first-order gradient-guided attacks like PGD and CW often get stuck at a local optima close to the starting point. Therefore, the loss value estimated by these attacks to solve the inner maximization problem might be significantly less than the true optimal value. Madry et al. try to mitigate such effects using a simple heuristic, i.e., picking a random starting point within the $L_\infty$ ball for each instance of the attack. However, local optima and true optimal values of the inner maximization problem can be very unevenly distributed, the chances of reaching a better solution by simply randomly restarting the gradient-guided attacks is very low. 

\begin{figure}[!hbt]
\centering
\vspace{-10pt}
\begin{subfigure}[t]{0.48\columnwidth}
	\centering
\includegraphics[width=\columnwidth]{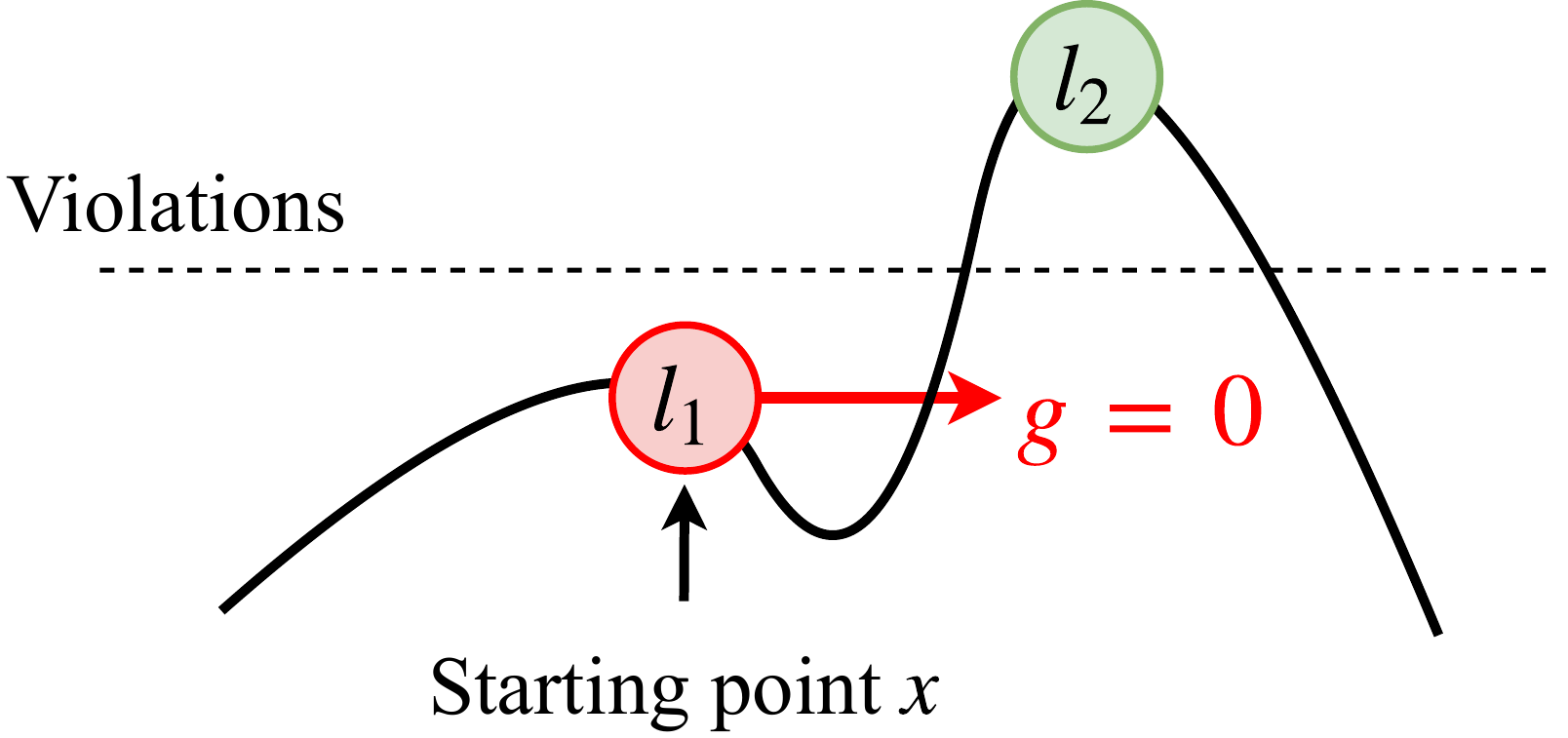}
\caption{PGD attack}
\label{subfig:iapgd}
\end{subfigure}
\centering
\begin{subfigure}[t]{0.48\columnwidth}
	\centering
\includegraphics[width=\columnwidth]{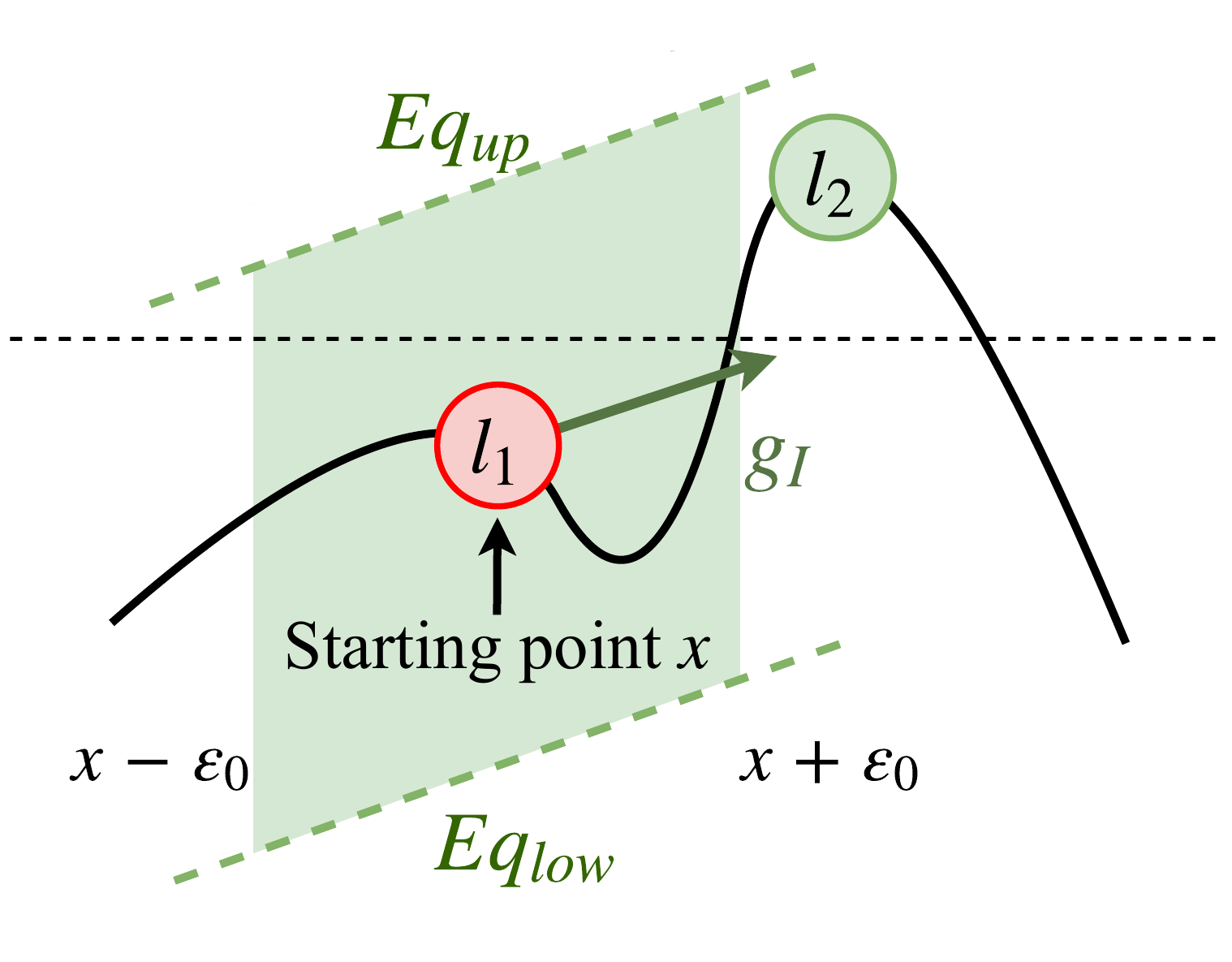}
\caption{Interval attack}
\label{subfig:ia}
\end{subfigure}

\caption{\bf \small The difference between regular gradient and interval gradient. Figure (a) illustrates that PGD attacks using regular gradients might get stuck at $l_1$. Figure (b) shows that interval gradient $g_I$ over an input region $B_{\epsilon_0}$ allows us to avoid such problem and successfully locate the violation $l_2$. Here $Eq_{up}$ and $Eq_{low}$ are the symbolic upper and lower bounds of the output as found by symbolic linear relaxation.}
\label{fig:ia}

\end{figure}

In this paper, we mitigate this problem, by using interval gradients instead of regular gradient to identify regions containing good starting points. 
Unlike regular gradients that summarize a function's behavior at a point, interval gradients are computed over a region and thus have a broader view of the functional landscape. Compared to more computationally expensive information like second-order gradients, the interval gradient is still cheap to compute and relies only on first-order information. Figure~\ref{subfig:iapgd} and Figure~\ref{subfig:ia} illustrate how interval gradient can identify the regions containing potential violations that regular point-gradient-based search from a random starting point cannot find. 

Our experiments confirm that the interval-gradient-guided search methods are more likely to find regions containing good starting points than random starts.  However, interval-gradient-guided search cannot directly identify concrete attack inputs. We address this by first using interval gradients to find input regions containing good starting points and then launch existing attacks like PGD within that region to find concrete violations. 

\noindent\textbf{Interval gradient ($g_I$)}. Sound over-approximation techniques like symbolic interval 
analysis provides the lower and upper bounds (in terms of two parallel linear equations over symbolic inputs) of 
a neural network's output for a bounded input region~\cite{reluval2018}. We define the interval gradient to be equal to the slope of these parallel lines. In this paper, we use \emph{symbolic linear relaxation}, which combines symbolic interval analysis with linear relaxation to provide tighter output bounds~\cite{shiqi2018efficient}. Essentially, such analysis produces two parallel linear equations ($Eq_{up}$ and $Eq_{low}$ as shown in Figure~\ref{fig:ia}) to tightly bound the output of each neuron.  For example, let us assume symbolic input relaxation provides the bounds of a neural network's output as $[2x,2x+3]$ after propagating input range $x=[0,1]$ through the network. Here, the interval gradient will be  $\frac{d([2x,2x+3])}{dx}=2$.

\noindent\textbf{Interval attack details.}
Algorithm~\ref{alg:ia} shows how we implemented the interval attack. The key challenge is to find a suitable value of $\epsilon_0$ representing the input region over which the interval gradient will be computed (Line 3 to Line 7)  at each iteration of interval-gradient-based search.
If $\epsilon_0$ is too large, the symbolic interval analysis will introduce large overestimation error leading to the wrong
direction. On the other hand, if $\epsilon_0$ is too small, the information from the
surrounding area might not be enough to make a clever update. Therefore, we 
dynamically adjust the size of $\epsilon_0$ during the attack. 

After a bounded number of iterations with interval gradients, we use PGD attack with a starting point from the region identified by the interval-gradient-based search to  locate a concrete violation as shown in Line \ref{alg:pgd_in_ia}.


\begin{algorithm}[!hbt]
	\caption{Interval Attack}
	\label{alg:ia}
	
	\begin{tabular}{|l|}
		\hline
		\textbf{Inputs}: $x$ $\leftarrow$ target input, $\epsilon$ $\leftarrow$ attack budget \ \\
		\textbf{Parameters}: \textbf{$t$}: iterations, \textbf{$\alpha$}: step size, \textbf{$\epsilon_0$}: starting point, \\ {p}: input region step size \\
		\textbf{Output}: $x'$ $\leftarrow$ perturbed $x$ \ \\
		\hline
	\end{tabular}
	\small
	\begin{spacing}{0.9}
		\begin{algorithmic}[1]
			\State $x' = x$
			\For {$i \gets 1$ to $t$}
				\State $\epsilon_0'=\epsilon_0$
				\State // Find smallest $\epsilon_0$ that might have violations
				\While {$argmax(d_{\epsilon_0}(x))==y$}
					\State $\epsilon_0'=\epsilon_0'*p$
					\If {$\epsilon_0'>=\epsilon/2$}  break // Prevent $\epsilon_0$ being too large
					\EndIf
				\EndWhile
				\State $g_I=\nabla_xL(d_{\epsilon_0}(x'),y)$ // From symbolic interval analysis
				\State $x' = x'+\alpha g_I$
				\State // Bound $x'$ within allowable input ranges
				\State $x' = clip(x', x-\epsilon, x+\epsilon)$   
				\If {$argmax(f_\theta(x'))\neq y$} \Return $x'$
				\EndIf
			\EndFor
			\State PGD Attack($x'$) \label{alg:pgd_in_ia}
			\State \Return $x'$
		\end{algorithmic}
	\end{spacing}
\end{algorithm}

\subsection{Effectiveness of Interval Attacks}
\label{sec:eval_ia}


We implemented symbolic interval analysis~\cite{shiqi2018efficient} and interval attacks
on top of \texttt{Tensorflow 1.9.0}\footnote{https://www.tensorflow.org/}.
All of our experiments are ran on a GeForce GTX 1080 Ti.
We evaluated the interval attack on two neural networks MNIST\_FC1 and MNIST\_FC2.
Both networks are trained to be adversarially robust using the Madry et al.'s technique~\cite{madry2017towards}. 
For solving the inner maximization problem while robust training, we used PGD attacks with 40 iterations and 0.01 as step size. We define the robustness region, $B_\epsilon(x)$, to be bounded by $L_\infty$ norm with $\epsilon=0.3$ over normalized inputs. The network details are shown in Table \ref{tab:ia_mnist}. We use the 1,024 randomly selected images from the MNIST test set to measure accuracy and robustness. MNIST\_FC1 contains two hidden layers each with 512 hidden nodes. and achieves 98.0\% test accuracy. Similarly, MNIST\_FC2 contains five hidden layers each with 2,048 hidden nodes and achieve 98.8\% test accuracy.

\begin{figure}[!hbt!]
	\centering
	\includegraphics[width=0.6\columnwidth]{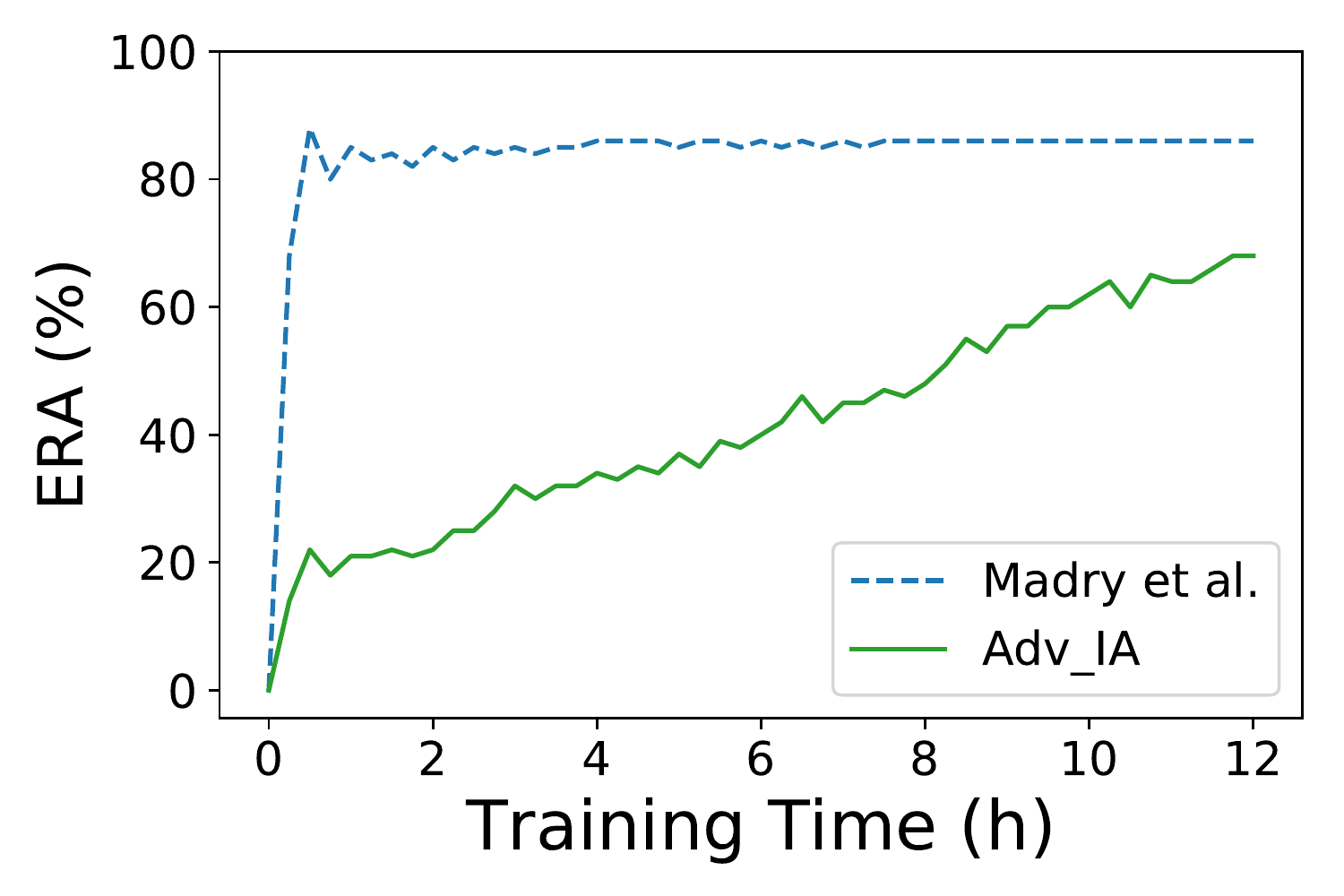}
	\caption{\bf \small Adversarially robust training using the interval attack does not converge well compared to using PGD attacks, given 12 hours of training time on the MNIST\_Small network with $L_\infty\leq 0.3$. The estimated robust accuracy of training using PGD attacks~\cite{madry2017towards} quickly converges to 89.3\% ERA, while training using the interval attack struggles to converge.}
	\label{fig:ia_adv}

\end{figure}



\begin{figure*}[!hbt]
	\begin{subfigure}[t]{\columnwidth}
		\includegraphics[width=\columnwidth]{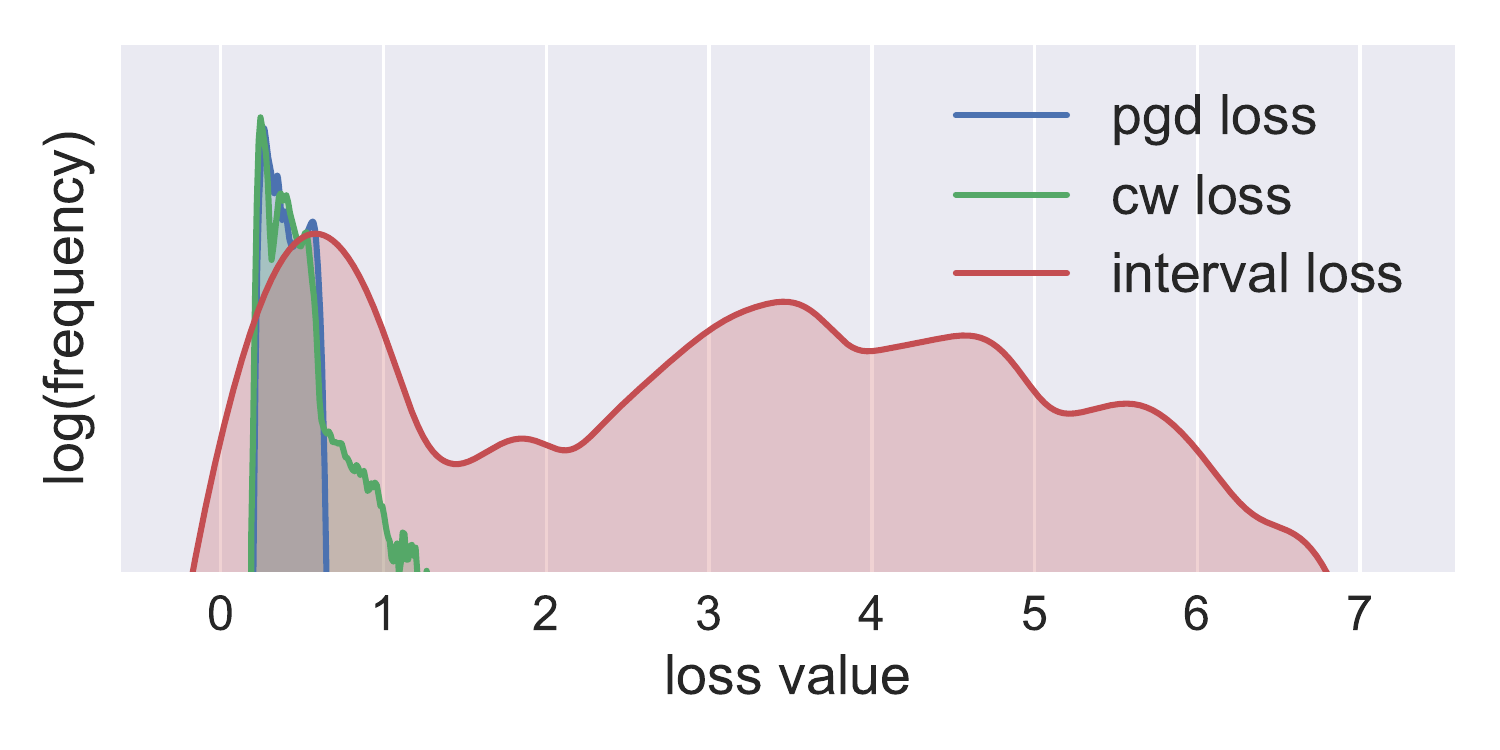}
		\caption{\bf \small MNIST image 12}
		\label{fig:iaimg1}
	\end{subfigure}
	\begin{subfigure}[t]{\columnwidth}
		\includegraphics[width=\columnwidth]{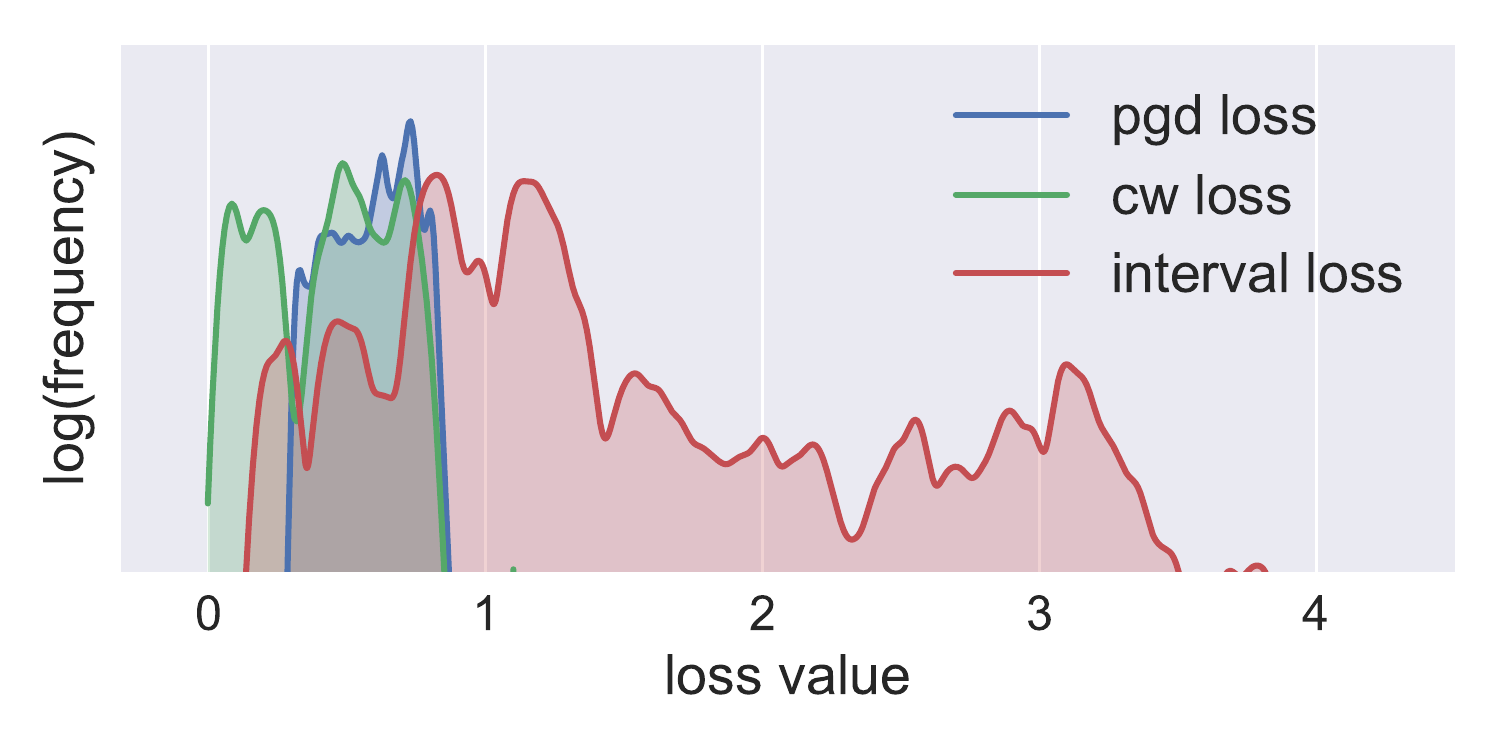}
		\caption{\bf \small MNIST image 31}
		\label{fig:iaimg2}
	\end{subfigure}
	\caption{\bf \small Distributions of the loss values from robustness violations found by PGD attacks (blue), CW attacks (green), and interval attacks (red) with 100,000 random starts within the allowable input range $B_\epsilon(x)$. The loss values found by CW and PGD attacks are very small and concentrated. However, interval attacks show there are still many distinct violations with much larger loss values.}
	\label{fig:iaimg}
\end{figure*}

\begin{table*}[!hbt]
\small
\centering
\begin{tabular}{|c|c|c|c|c|c|c|}
\hline
\multirow{2}{*}{Network} & \multirow{2}{*}{\# Hidden units} & \multirow{2}{*}{\# Parameters} & \multirow{2}{*}{ACC (\%)} & \multicolumn{3}{c|}{Attack success rate (\%)} \\ \cline{5-7} 
&                               &                             &                               & PGD       & CW        & Interval attack     \\ \hline
MNIST\_FC1                    & 1,024                         & 668,672                     & 98.1        & 39.2      & 42.2  (1.08$\times$)    &\textbf{56.2} (1.43$\times$)           \\ \hline
MNIST\_FC2                    & 10,240                        & 18,403,328               & 98.8        & 34.4      & 32.2 (0.94$\times$)    & \textbf{44.4} (1.38$\times$)             \\ \hline
\end{tabular}
\caption{\bf \small Attack success rates of PGD, CW and interval attacks ($\epsilon=0.3$) on MNIST\_FC1 and MNIST\_FC2 networks. The highest success rates are in bold. Both networks are adversarially robust trained using PGD attacks with $\epsilon=0.3$~\cite{madry2017towards}.}
\label{tab:ia_mnist}
\vspace{-10pt}
\end{table*}

\noindent\textbf{Attack effectiveness.} The interval attack is very effective at finding violations in both of our tested 
networks that are relatively robust against PGD and CW
attacks with less than 35\% success rate,.
We compared the strength of the interval attack against the state-of-the-art PGD and CW attacks
given the same amount of attack time, as shown in Table~\ref{tab:ia_mnist}.
The interval attack is able to achieve up to 40\% increase in the attack success rate of PGD and CW.
The iterations for PGD and CW are around 7,200 for MNIST\_FC1 and 42,000 for MNIST\_FC2.
We ran the interval attack for 20 iterations against MNIST\_FC1 and MNIST\_FC2,
which took 53.9 seconds and 466.5 seconds respectively.

Our results show that using only first-order gradient information,
the interval attack can find significantly more robustness violations than PGD and CW attacks in
adversarially robust networks.

\noindent\textbf{Inefficiency of adversarially robust training with interval attacks.}
One obvious way of increasing the robustness of adversarially robust trained networks against interval-based attacks is to 
use such attacks to solve the inner maximization problem during training. However, due to the highly uneven distribution of the loss values (as shown in Figure~\ref{fig:iaimg}), robust training with interval attacks often struggle to converge. 
To demonstrate that, we evaluated adversarial robust training using interval attacks on MNIST\_Small network for $L_\infty\leq 0.3$. As shown in the Figure \ref{fig:ia_adv}, even after 12 hours training time for such a small network, the 
interval-based adversarially robust training does not converge as well as its PGD-based counterpart. Therefore, instead of improving the adversarially robust training schemes, we focus on scaling verifiably robust training, which can also defend against stronger attackers. In Section~\ref{sec:eval_train_schemes}, we show that \sys can achieve similar test accuracy with up to 95.2\%  more verified robust accuracy than existing adversarially robust training schemes, using $3\times$ less training time.

\section{Methodology}
\label{sec:methodology}

In this section, we discuss the design of \sys in detail. \sys is a generic defense that can work with any verifiably robust training methods to efficiently and scalably train robust neural networks.
Specifically, we propose two key techniques, \emph{\SOA} and \emph{\DMT}. \Soa significantly cuts down the overhead of \soov with stochastic sampling.  Similarly, \dmt helps the trained network to simultaneously achieve both high VRA and high test accuracy.
 
\subsection{Motivation}
\label{subset:problem}

\begin{figure}[!hbt]
	\begin{subfigure}[t]{0.48\columnwidth}
		\includegraphics[width=\columnwidth]{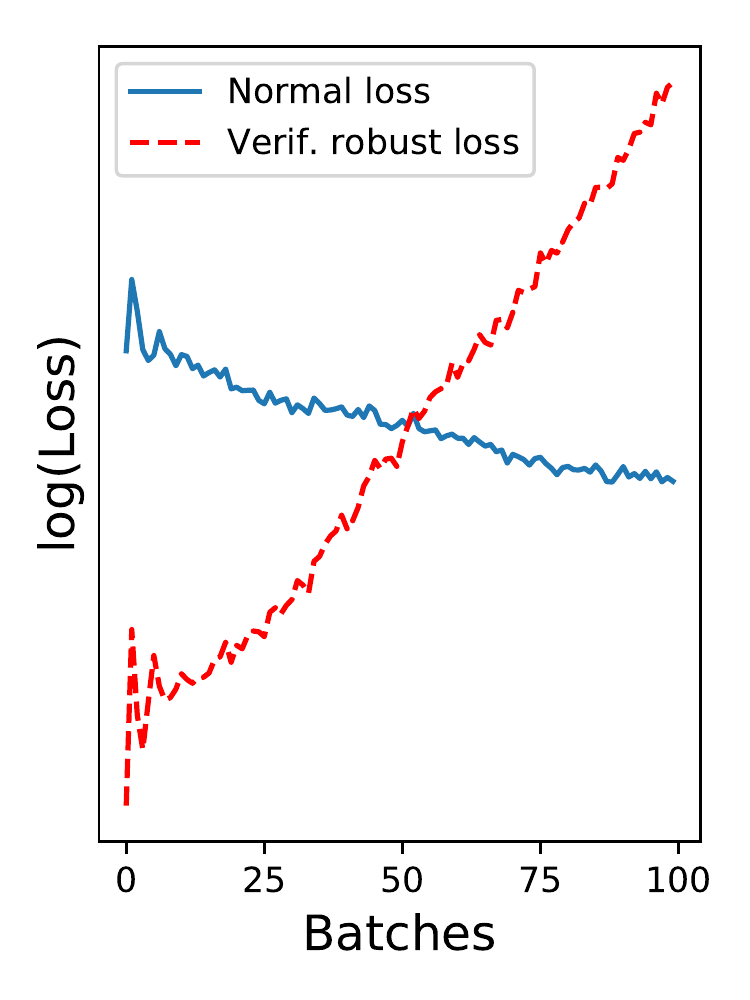}
		\caption{\bf \small Regular training}
		\label{fig:conflict_n}
	\end{subfigure}
	\begin{subfigure}[t]{0.48\columnwidth}
		\includegraphics[width=\columnwidth]{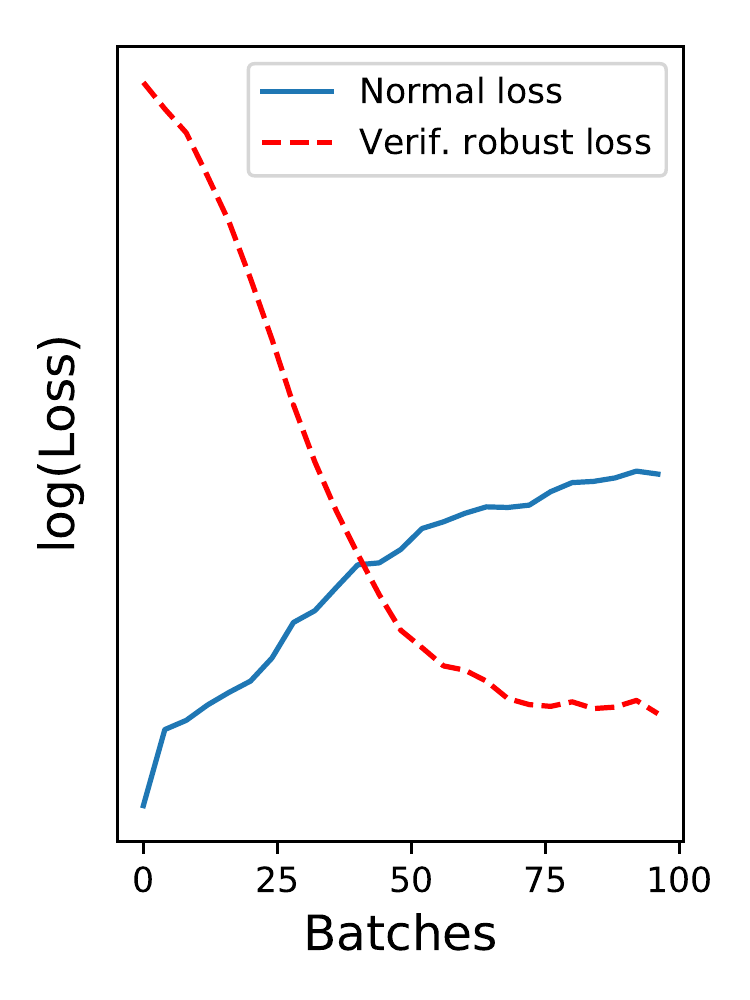}
		\caption{\bf \small Robust training}
		\label{fig:conflict_r}
	\end{subfigure}
	\caption{\bf \small The conflicting changes in regular loss and verifiable robust loss while training two CIFAR\_Small networks. The left one is regular training after x epochs of verifiably robust training (Wong et al.'s method~\cite{wong2018scaling}) and the right one is verifiably robust training after x epochs of regular training.}
	\vspace{-15pt}
\end{figure}

Our design decisions for \sys are based on the observation that existing verifiably robust training methods suffer from two main scalability issues described below.

\noindent\textbf{High computation and memory cost.} Verifiably robust training
uses sound over-approximation techniques like symbolic interval analysis, abstract interpretation, convex polytope etc. to solve the inner maximization problem as mentioned in Section~\ref{sec:background}. All existing verifiably robust training schemes compute sound over-approximations of the network robustness for all training samples. This causes significant overhead both in terms of computation and memory. The state-of-the-art sound over-approximation techniques, despite impressive progress, are still hundreds of times slower and require thousands of times more memory than the regular forward propagation in
a neural network~\cite{wong2018provable,wong2018scaling,weng2018towards,shiqi2018efficient}. Most of the computational overhead in sound over-approximation techniques results from computing the precise bounds on the outputs of different hidden nodes of a network. Similarly, the amount of memory required for sound over-approximation increases significantly with the number of hidden nodes in a network. For instance,
for verifiably robust training of a CIFAR residual network with around 32,000 hidden nodes requires up to  
around 300GB of memory while the regular training only needs 7MB. 

\noindent\textbf{Conflict between verifiable robustness and accuracy.}\
We observe that verifiable robustness and test accuracy start conflicting after the initial training epochs, i.e., increasing one often decreases the other. Figure~\ref{fig:conflict_n} and Figure~\ref{fig:conflict_r} clearly shows this effect. For the first experiment, we start with a CIFAR\_Small
network pre-trained for 60 epochs with Wong et al.'s verifiably robust training method~\cite{wong2018scaling}. 
When we further trained this network with regular training, as shown in Figure \ref{fig:conflict_n}, the normal loss decreases (i.e., the test accuracy improves) but the verifiable robust loss increases (i.e., the VRA decreases).
For the second experiment, we started with CIFAR\_Small network pre-trained with regular training after 60 epochs. Training this network with verifiable robust training increases VRA but decreases test accuracy 
as shown by the loss values in Figure~\ref{fig:conflict_r}.
The results demonstrate that improving verifiable robustness and test accuracy becomes conflicting after the initial training epochs.

\subsection{\SOA}
We propose \emph{\soa} to reduce the computational and memory cost of the verifiable robust training process.
The key insight behind our approach we should try to minimize the usage of expensive sound over-approximation methods without affecting the verified robustness of the final network. We observe that one does not actually need sound over-approximations of robustness violations over the entire training dataset during all intermediate steps of the training process. The verifiably robust training can efficiently minimize the verifiable robust loss as long as it is guided with sound over-approximation of robustness violations over a representative subset of the training data. Therefore, we use random subsampling to pick a representative subset of training data and use sound over-approximation process to estimate robustness violations over this subset. This step significantly speeds up the training process.
    
\Soa randomly samples $k$ training data points as $\mathcal{D}_k$ from  $\mathcal{D}_0$, i.e., $n$ training data points. We find that the verifiable robust loss values computed over $\mathcal{D}_k$ are representative of the verifiable robust loss values computed over the entire training dataset $\mathcal{D}_0$.

Formally, let the $\mathcal{D}_k$ denote the representative $k$ data samples
out of the entire training set $\mathcal{D}_0$. Then, the new inner maximization problem
for \soa can be defined as:

\begin{equation}
\max_{\tilde{x}\in B_\epsilon(x)} \E_{(x,y)\sim\mathcal{D}_k}L(d_\epsilon(x), y)
\label{eq:robust_term}
\end{equation}

\begin{figure}[bt!]
\vspace{-10pt}
	\includegraphics[width=\columnwidth]{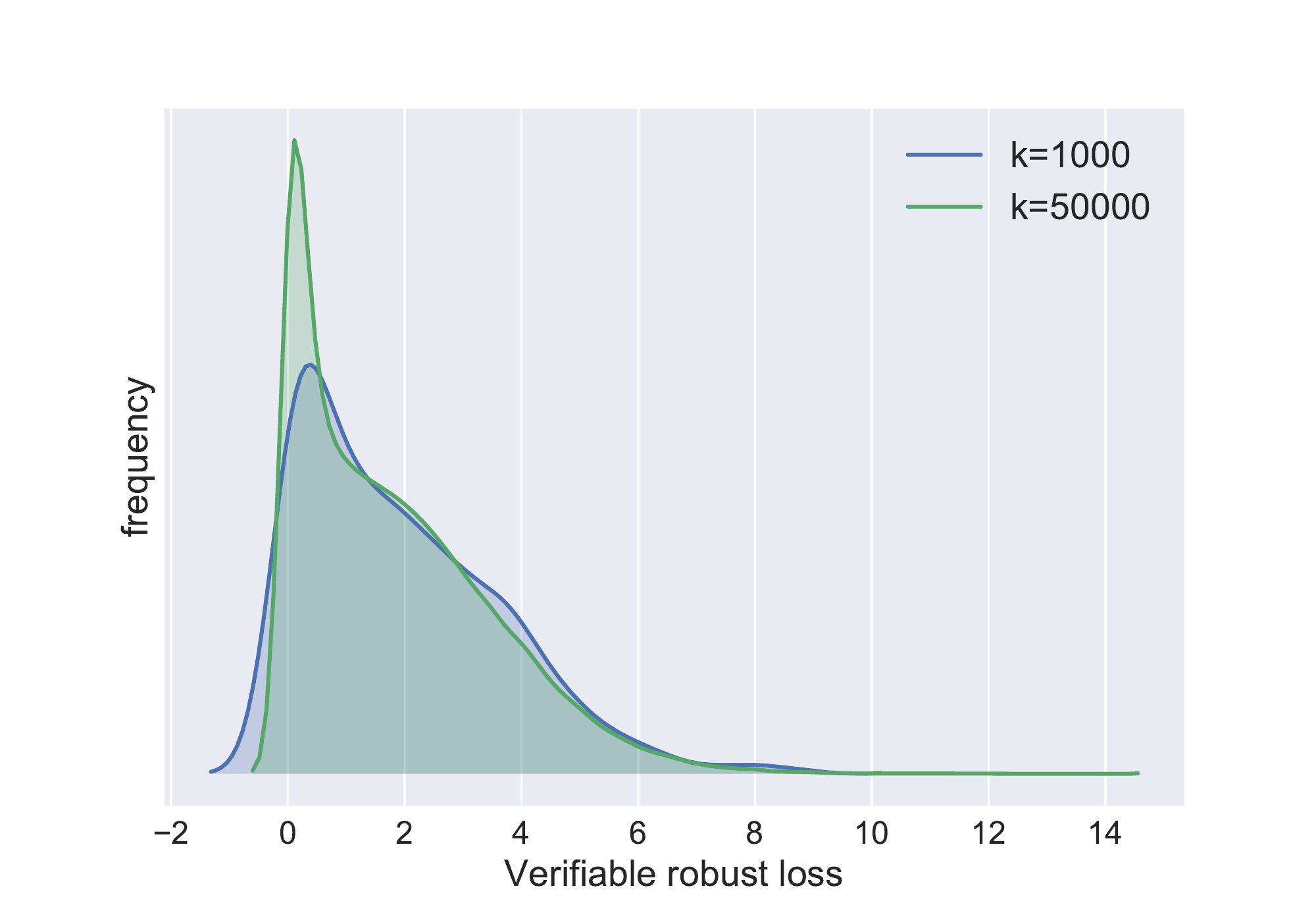}
    \caption{\bf \small The distributions of verifiable robust loss from the entire training set $\mathcal{D}_0$ and from the sampled training set $\mathcal{D}_k$ (k=1000) are very similar.}
	\label{fig:sample}
\end{figure}

To test the effectiveness of subsampling to estimate robust loss over the entire training dataset, we randomly subsamples 1,000 data points ($k=1,000$) out of 50,000 in CIFAR training dataset. We compute the verifiable robust loss (i.e., estimated verifiable robustness violations) over both the subsampled data points and the entire training dataset. Figure \ref{fig:sample} clearly shows that the distribution of verifiable robust loss
$\mathcal{D}_k$ over the sub-sampled set is very close to the loss distribution over the entire training dataset $\mathcal{D}_0$. The two distributions significantly overlap and the verifiable robust loss values are between 0 to 2 for most of the data points. 

Figure \ref{fig:sample} clearly demonstrates that we can accurately estimate the verifiable robust loss values
by computing sound over-approximation of the robustness vioaltions over a randomly subsampled training dataset.
This significantly reduces the time and memory requirements for \sys. For instance, by sampling $k=1,000$ training data points out of $50,000$, the computational cost of verifiably robust training gets down to 2\% of the original cost using the entire training set.

Such finding is not surprising given that stochastic sampling is a common technique used in Machine Learning~\cite{anthony2009neural}. As we discussed in Section \ref{sec:background}, all training procedures for neural networks use sampling to perform the Empirical Risk Minimization (ERM) process for estimating the loss values over the underlying true distribution. Specifically, the $n$ data points in the training dataset are assumed to be sampled from the
underlying distribution $\mathcal{D}$ and are treated as an empirical distribution $\mathcal{D}_0$. \Soa simply takes such procedures one step further by subsampling the empirical distribution $\mathcal{D}_0$ for computing verifiable robust loss.

\noindent\textbf{Integration with batch training.}
We further utilize the randomness in the batch generation process to integrate \soa with batch training that is commonly used for neural networks. Essentially, we pick $k$ samples from different batches. Let $m$ and $n$ denote the batch size and the number of original training data points. Here, the training process will use $\frac{n}{m}$ batches. By distributing $k$ random samples over these batches, we get $k' = \frac{k}{n/m}$ samples
per batch. Before updating the weights of the neural network for each batch through back propagation, we compute the estimated verifiable robust loss from $k'$
random samples within the batch. For instance, in Figure~\ref{fig:sample}, given batch size $m=50$
and training dataset size $n=50,000$, randomly sampling $k=1,000$ from the training set is equivalent to
randomly picking $k'=1$ per batch. Our experimental results in Section~\ref{sec:eval} shows that setting $k'=1$ achieves high VRA while running up to $14\times$ faster than the state-of-the-art verifiably robust training
schemes~\cite{wong2018scaling}. \soa can be customized by changing the value of $k$ as a hyperparameter.

\subsection{Dynamic Mixed Training}
\label{subsec:dmt}
Our second technique solves the conflict between the test accuracy and the verifiable robustness.
Most state-of-the-art robust training schemes struggle to efficiently achieve both high test accuracy and high verifiable robustness~\cite{dvijotham2018training,mirman2018differentiable, wong2018scaling}. To simultaneously increase both the test accuracy and the verifiable robustness, we dynamically
balance the goals of training for each epoch using a dynamic loss function.


Specifically, we define our dynamic loss function as the following:

\begin{equation}
L_{mixed} = (1-\alpha)\E_{(x,y)\sim\mathcal{D}_0}L(f_\theta(x), y)+\alpha\E_{(x,y)\sim\mathcal{D}_k}L(d_\epsilon(x), y)
\label{eq:mixed_loss}
\end{equation}

$L_{mixed}$ is a weighted sum of the expectation of both
the regular loss $\E_{(x,y)\sim\mathcal{D}_0}L(f_\theta(x), y)$
and the verifiable robust loss $\E_{(x,y)\sim\mathcal{D}_k}L(d_\epsilon(x), y)$.
The hyperparameter $\alpha$,  which takes a value between 0 and 1, biases the dynamic loss value towards either regular loss
or verifiably robust loss.
When $\alpha$ is large, the training process tends to find the weights $\theta$ that minimize
the verifiable robust loss, increasing VRA. In contrast, if $\alpha$ is small, Eqn.~\ref{eq:mixed_loss}
emphasizes more on the regular loss and thus enhances the test accuracy.

\noindent\textbf{Dynamic Loss.}
We dynamically adjust $\alpha$ per epoch to prioritize different emphasis on the regular
loss and the verifiably robust loss.  The dynamic loss function allows us to adaptively maximize the test accuracy and VRA simultaneously based on the the current test accuracy. It can balance the gradients generated from the loss of
both the regular and the verifiably robust training tasks.
Therefore, we can avoid the training being completely dominated by
the one with a larger gradient.
Using dynamic loss, we are able to robustly train the networks that have up to 13.2\% higher test accuracy compared to the state-of-the-art verifiable robust training schemes~\cite{wong2018scaling} with similar VRA (Section \ref{sec:eval}).
Besides the test accuracy improvement, another significant benefit of \sys is that it helps convergence of the training process especially when either the regular loss or verifiable robust loss functions are difficult to optimize by themselves.

In our experiments, we use a specific $\alpha$
function as defined in Section~\ref{sec:eval_train_schemes}. In general, $\alpha$ can be any reasonably smooth function. For example, grid search can be used to find the best $\alpha$
for each epoch. Alternatively, the best $\alpha$ for different epochs and different networks can potentially be policies learned using reinforcement learning techniques. We leave exploring these directions as future work.



\section{Evaluation}
\label{sec:eval}

We implemented \soa and \dmt as part of \sys, built on \texttt{PyTorch 0.4.0}\footnote{https://pytorch.org/}.
In \sys, we used symbolic linear relaxation as the sound over-approximation method proposed by Wang et al.~\cite{shiqi2018efficient}.
In this section, we evaluate \sys on 6 different networks trained over 3 different datasets and compare \sys's performance against 4 state-of-the-art verifiably and adversarially robust training methods. We compare these robust training techniques against \sys using three metrics: test accuracy (ACC), estimated robust accuracy under PGD attacks (ERA), and verified robust accuracy (VRA). Unless otherwise specified,  we use ERA to denote the robust accuracy as measured with PGD attacks for the rest of the paper. We summarize our main results below:
\begin{enumerate}
\item Compared to the state-of-the-art adversarially robust training methods~\cite{madry2017towards, sinha2018certifying} that have zero VRA, \sys can obtain up to 95.2\% VRA with similar ERA(PGD) and ACC, using 3 times less training time on average.
\item Compared to state-of-the-art verifiably robust training methods~\cite{wong2018scaling, mirman2018differentiable}, \sys can achieve up to 13.2\% higher ACC and 6.5\% higher VRA, allowing verifiably robust networks to be trained for real-world applications . 
\item \sys requires, on average, 10 times less training time compared to state-of-the-art verifiably robust training schemes~\cite{wong2018scaling} and thus can easily scale to large network trained on the ImageNet-200 dataset. For such large network, \sys outperforms the existing verifiably robust training methods by up to 14.3\% VRA within 12-hour of training time.
\end{enumerate}

\subsection{Experimental Setup}
\label{subsec:setup}


\begin{table}[!hbt]
	\small
	\begin{tabular}{|c|c|c|c|}
		\hline
		Network         & Type     & \# hidden units & \# parameters \\ \hline
		MNIST\_Small  & Conv     & 4,804            & 166,406        \\ \hline
		MNIST\_large  & Conv     & 28,064           & 1,974,762       \\ \hline
		CIFAR\_Small  & Conv     & 6,244            & 214,918        \\ \hline
		CIFAR\_Large  & Conv     & 62,464           & 2,466,858       \\ \hline
		CIFAR\_Resnet & Residual & 31,720          & 1,145,410       \\ \hline
		ImageNet\_Resnet & Residual & 172,544 & 2,310,664 \\ \hline
	\end{tabular}
	\caption{\bf \small Architectures of the 6 neural networks used for evaluating \sys.}
	\label{tab:networks}
\end{table}

\noindent\textbf{Datasets \& networks.} We evaluate \sys on three different datasets: MNIST digit classification~\cite{lecun1998gradient}, CIFAR10 image classification~\cite{krizhevsky2009learning} and ImageNet-200\footnote{Imagenet-200 is a subset of the ImageNet dataset with 200 classes instead of 1,000. Also, each piece of training data is cropped to be 3$\times$64$\times$64.}~\cite{tiny-imagenet}. We use six different network architectures for the experiments. The details of the networks are shown in Table \ref{tab:networks}. We refer the interested readers to Appendix \ref{sec:ans} for further details.

Unless otherwise specified, similar to Wong et al.~\cite{wong2018scaling}, we train 60 epochs in total with batch size of 50 for all experiments. We follow the standard normalization procedures (described below before training for all of the datasets~\cite{wong2018scaling,dvijotham2018training,madry2017towards,sinha2018certifying}. For the MNIST dataset, we scale the inputs to be within $[0,1]$ with $\mu=0.5$. We use the Adam optimizer~\cite{kingma2014adam} with learning rate 0.001 and decay it by a factor of 0.6 for every 5 epochs. We use $\epsilon$ starting from 0.01 to the values used for measuring VRA over the first 10 epochs\footnote{Since sound over-approximations over a network during the early stages of training is more expensive than later stages, decaying $\epsilon$ is necessary to bootstrap the training process.}. Similarly, for the CIFAR dataset, we normalize inputs with $\mu=[0.485, 0.456, 0.406],
\sigma=[0.225, 0.225, 0.225]$. We use the SGD optimizer~\cite{robbins1985stochastic,kiefer1952stochastic} with learning rate 0.05 and decay it by 0.6 for every 5 epochs. We use $\epsilon$ starting from 0.001 to values used for measuring VRA over the first 10 epochs. We use the same configurations for ImageNet-200 dataset too. All of our experiments are ran on a GeForce GTX 1080 Ti. For all networks trained using the same dataset, we use the configurations as well as the same randomly selected set of test images to compute ACC, ERA, and VRA.

\label{subsubsec:alpha}

\noindent\textbf{Dynamic $\alpha$ function of \sys.} We used the following $\alpha$ function for \sys's dynamic loss function:

\[
\alpha_t= 
\begin{cases}
\alpha_0, & \text{if } t = 0 \\
\alpha_{t-1} + 0.05,  & \text{if $acc > acc_{0}$  } \\
\alpha_{t-1} - 0.05,  & \text{if $acc \leq acc_{0}$}
\end{cases}
\]

Here $acc$ denotes the accuracy of a sub-sampled training dataset for the current epoch $t$. $acc_{0}$ and $\alpha_0$ denote the desired  accuracy and the initial $\alpha$ value, respectively.
In this conditional function, $\alpha$ takes an initial value of $\alpha_0$ at the beginning (i.e., $t = 0$) of the
batch training process. Afterwards, for each epoch, $\alpha$ changes
according to the accuracy of current epoch. When the test accuracy increases and becomes larger than targeted $acc_{0}$, we increase $\alpha$ by $0.05$. Similarly, if the test accuracy decreases and becomes less than $acc_{0}$, we deduct $0.05$ from the current value of $\alpha$. 

Since the MNIST networks, unlike the other networks we tested, can be easily trained to have high test accuracy, we want \sys to spend more effort on improving verifiable robustness in \dmt. Therefore, we set $\alpha_0$ to be $0.8$.
By contrast, for CIFAR networks, where achieving high accuracy is hard, we set $\alpha_0$ to $0.5$.

\begin{table*}[!hbt]
\centering
\small
\begin{tabular}{cclrrrrr}
\hline
\multicolumn{1}{l}{Network}    & \multicolumn{1}{c}{Epsilon} & \multicolumn{1}{c}{Method} & \multicolumn{1}{c}{Batch Time (s)} & \multicolumn{1}{c}{Training Time} & \multicolumn{1}{c}{ACC (\%)} & \multicolumn{1}{c}{ERA (\%)} & \multicolumn{1}{c}{VRA (\%)} \\ \hline
\multirow{7}{*}{MNIST\_Small} & \multirow{4}{*}{0.1}         & Regular training                    & 0.006                               & 7m12s                             & 98.6                               & 0.8                                & 0                             \\  \cline{3-8} 
&                              & Madry et al.~\cite{madry2017towards}                & 0.027                               & 32m4s                            & {99.1}                     & 96.0                              & 0                             \\  
&                              & Sinha et al.~\cite{sinha2018certifying}                & 0.026                               & 31m12s                            & 98.7                              & 58.5                              & 0                             \\  
&                              & \sys                        & \textbf{0.011}                      & \textbf{13m12s}                    & 98.5                               & {96.3}                               & \textbf{91.6}                          \\ \cline{2-8} 
& \multirow{3}{*}{0.3}         & Madry et al.~\cite{madry2017towards}                & 0.027                               & 32m24s                            & 98.8                               & {89.3}                      & 0                             \\  
&                              & Sinha et al.~\cite{sinha2018certifying}                & 0.026                               & 31m12s                            & {98.9}                     & 0                                  & 0                             \\  
&                              & \sys                         & \textbf{0.011}                      & \textbf{13m48s}                    & 95.4                               & 87.3                               & \textbf{52}                            \\ \hline
\multirow{7}{*}{MNIST\_Large} & \multirow{4}{*}{0.1}         & Regular training                    & 0.007                               & 8m24s                             & 99.3                               & 26.8                               & 0                             \\ \cline{3-8} 
&                              & Madry et al.~\cite{madry2017towards}                & 0.063                               & 1h15m                            & 98.9                              & 96.7                              & 0                             \\  
&                              & Sinha et al.~\cite{sinha2018certifying}                & 0.063                               & 1h16m                            & 99.0                              & 70.1                              & 0                             \\  
&                              & \sys                         & \textbf{0.051}                      & \textbf{1h1m}                   & {99.5}                      & {98.2}                      & \textbf{95.2}                   \\ \cline{2-8} 
& \multirow{3}{*}{0.3}         & Madry et al.~\cite{madry2017towards}                & 0.063                               & 1h15m                            & \multicolumn{3}{c}{Does not converge}                                                                       \\  
&                              & Sinha et al.~\cite{sinha2018certifying}                & 0.066                               & 1h19m                            & {99.1}                              & 0.1                               & 0                             \\    
&                              & \sys                         & \textbf{0.054}                      & \textbf{1h5m}                   & 96.6                      & {89.4}                      & \textbf{58.4}                 \\ \hline

\multirow{4}{*}{CIFAR\_Small}  & \multirow{4}{*}{0.0348}     & Regular training                   & 0.011                              & 10m57s                                  & 77.3                                & 18.6                              & 0                            \\ \cline{3-8} 
&                             & Madry et al.~\cite{madry2017towards}               & 0.194                              & 3h15m                                & 71.2                             & 54.3                             & 0                            \\
&                             & Sinha et al.~\cite{sinha2018certifying}               & 0.192                              & 3h11m                                & {71.8}                    & 11.3                             & 0                            \\
&                             & \sys        & \textbf{0.015}                              & \textbf{14m52s}                                  & 71.1                                & {54.6}                     & \textbf{37.6}                         \\ \hline
\multirow{4}{*}{CIFAR\_Large}  & \multirow{4}{*}{0.0348}     & Regular training                   & 0.012                              & 12m3s                                  & 86.4                              & 14.3                              & 0                            \\ \cline{3-8} 
&                             & Madry et al.~\cite{madry2017towards}               & 0.289                              & 4h49m                                & {80.9}                    & {63.6}                    & 0                            \\
&                             & Sinha et al.~\cite{sinha2018certifying}               & \textbf{0.204}                              & \textbf{3h24m}                                & 75.1                             & 13.7                             & 0                            \\
&                             & \sys        & 0.239                               & 3h59m                                & 77.9                              & 63.5                              & \textbf{41.6}                \\ \hline
\multirow{4}{*}{CIFAR\_Resnet} & \multirow{4}{*}{0.0348}     & Regular training                   & 0.014                              & 14m39s                                  & 84.2                                & 22.6                              & 0                            \\ \cline{3-8} 
&                             & Madry et al.~\cite{madry2017towards}               & 0.306                              & 5h6m                                & \multicolumn{3}{c}{Does not converge}                                                                     \\
&                             & Sinha et al.~\cite{sinha2018certifying}               & 0.212                              & 3h32m                                & 77.3                             & 17.7                              & 0                            \\
&                             & \sys        & \textbf{0.109}                              & \textbf{1h49m}                                 & {77.5}                     & {61.0}                     & \textbf{35.5}                  \\ \hline 
\end{tabular}
\caption{\bf \small Comparison between \sys and existing state-of-the-art adversarially robust training methods. The best robust training numbers are shown in bold. Compared to the adversarially robust trained networks, after training 60 epochs, \sys with k=1 can provide similar test accuracy and ERA with up to 95.2\% more verifiable robustness within shorter training time.}
\label{tab:main1}
\vspace{-10pt}
\end{table*}

\begin{table*}[]
	\small
	\centering
	\begin{tabular}{cclrrrrr}
		\hline
		{Network}    & \multicolumn{1}{l}{$\epsilon$} & \multicolumn{1}{l}{Method} & \multicolumn{1}{l}{Batch Time (s)} & \multicolumn{1}{l}{Training Time} & \multicolumn{1}{l}{ACC (\%)} & \multicolumn{1}{l}{ERA (\%)} & {VRA (\%)} \\ \hline
		\multirow{6}{*}{MNIST\_Small} & \multirow{3}{*}{0.1}   & DIFFAI~\cite{mirman2018differentiable}                      & 0.013                               & 12h                             & 97.8                               & 93.1                               & 60.6                          \\  
		&                              & Wong et al.~\cite{wong2018scaling}               & 0.078                                & 12h                           & 98.6                               & {96.9}                        & {95.7}                 \\  
		&                              & \sys                         & {0.026}                      & 12h                    & \textbf{99.5}                               & 98.0                               & \textbf{97.1}                          \\ \cline{2-8} 
		& \multirow{3}{*}{0.3}         & DIFFAI~\cite{mirman2018differentiable}                      & 0.012                              & 12h                             & {98.2}                               & 82.9                               & 0                             \\  
		&                              & Wong et al.~\cite{wong2018scaling}               & 0.079                               & 12h                           & 91.2                               & 83.1                               & {57.8}                 \\  
		&                              & \sys                          & {0.028}                      & 12h                    & \textbf{94.0}                               & 86.1                               & \textbf{60.1}                            \\ \hline
		\multirow{6}{*}{MNIST\_Large} & \multirow{3}{*}{0.1}         & DIFFAI~\cite{mirman2018differentiable}                      & 0.022 & 12h & \multicolumn{3}{c}{Does not converge}                                             \\  
		&                              & Wong et al.~\cite{wong2018scaling}               & 0.687 & 12h & 99.2 & 98.3 & \textbf{96.4}                           \\  
		&                              & \sys                          & {0.051}                      & 12h                   & \textbf{99.5}                      & {98.2}                      & {95.2}                   \\ \cline{2-8} 
		& \multirow{3}{*}{0.3}         & DIFFAI~\cite{mirman2018differentiable}                      & 0.023 & 12h & \multicolumn{3}{c}{Does not converge}                                                                   \\  
		&                              & Wong et al.~\cite{wong2018scaling}               & 0.689 & 12h & 86.6 & 71.7 & 53.1                            \\  
		&                              & \sys                        & {0.054}                      & 12h                   & \textbf{96.6}                      & {89.4}                      & \textbf{58.4}                 \\ \hline
		
		\multirow{3}{*}{CIFAR\_Small}  & \multirow{3}{*}{0.0348}     & DIFFAI~\cite{mirman2018differentiable}                     & 0.015                     & 12h                         & 66.3                              & 41.9                              & 4.0                          \\
		&                             & Wong et al.~\cite{wong2018scaling}                & 0.248                              & 12h                                & 62.5                              & 52.3                              & {47.5}                \\
		&                             & \sys       & {0.087}                              & 12h                                  & \textbf{69.4}                                & {56.5}                     & \textbf{48.2}                        \\ \hline
		\multirow{3}{*}{CIFAR\_Large}  & \multirow{3}{*}{0.0348}     & DIFFAI~\cite{mirman2018differentiable}                     & 0.025 & 12h &   \multicolumn{3}{c}{Does not converge}                                      \\
		&                             & Wong et al.~\cite{wong2018scaling}                & 2.157 & 12h & 61.9 & 56.4 & 49.4                     \\
		&                             & \sys         & {0.268}                               & {12h}                                & \textbf{74.1}                              & 62.7                              & \textbf{50.2}                \\ \hline
		\multirow{3}{*}{CIFAR\_Resnet} & \multirow{3}{*}{0.0348}    & DIFFAI~\cite{mirman2018differentiable}                    & {0.035} & 12h &75.6  & 35.8 & 0.0                            \\
		&                             & Wong et al.~\cite{wong2018scaling}                & 1.914 & 12h & 58.1 & 51.7 & 43.9                    \\
		&                             & \sys         & {0.281}                               & {12h}                                & \textbf{71.3}                              & 60.5                              & \textbf{50.4}                  \\ \hline 
	\end{tabular}
	\caption{\bf \small Comparison between state-of-the-art verifiably robust training methods and \sys. The best numbers are shown in bold. With 12 hours of training time, compared to the verifiably robust trained networks, \sys can achieve up to 13.2\% ACC improvements and 6.5\% VRA improvements.}
	\label{tab:main2}
\end{table*}

\begin{table}[]
	\small	
	\begin{tabular}{c@{\hskip5pt}c@{\hskip7pt}l@{\hskip3pt}r@{\hskip5pt}r@{\hskip8pt}r}
		\hline
		\multicolumn{1}{l}{Network}                                                 & $\epsilon$                 & Method                                                         & \multicolumn{1}{c}{\begin{tabular}[r]{@{}r@{}}Training\\Time\end{tabular}} & \begin{tabular}[r]{@{}r@{}}Others\\\hline\sys \end{tabular}  & \begin{tabular}[c]{@{}c@{}}ACC\\(\%)\end{tabular} \\ \hline
		\multirow{6}{*}{\begin{tabular}[c]{@{}c@{}}MNIST\\ \_Small\end{tabular}} & \multirow{3}{*}{0.1}    & DIFFAI~\cite{mirman2018differentiable}                                                     & \textgreater{}24h       & \textgreater{36$\times$}                                                            & N/A                                              \\
		&                         & Wong et al.~\cite{wong2018scaling}                                                    & \textgreater{24h}                                & \textgreater{36$\times$}                                   & N/A                                               \\
		&                         & \sys                                                                  & {39m26s}  & {-}                                       & 99.5                                               \\ \cline{2-6} 
		& \multirow{3}{*}{0.3}    & DIFFAI~\cite{mirman2018differentiable}                                                   & \textgreater{}24h                         &\textgreater{28$\times$}                                          & N/A                                               \\
		&                         & Wong et al.~\cite{wong2018scaling}                                                           & \textgreater{}24h          & \textgreater{28$\times$}                                                          & N/A                                              \\
		&                         & \sys                                                                   & {50m12s}                     &-                     & 94.0                                               \\ \hline
		\multirow{6}{*}{\begin{tabular}[c]{@{}c@{}}MNIST\\ \_Large\end{tabular}} & \multirow{3}{*}{0.1}    & DIFFAI~\cite{mirman2018differentiable}                                                      & \textgreater{24h} & \textgreater{26$\times$} & N/A                                       \\
		&                         & Wong et al.~\cite{wong2018scaling}                                                        & 3h59m    & 4$\times$ & 98.9                               \\
		&                         & \sys                                                                 & {55m12s}                          & -              & {99.5}                     \\ \cline{2-6} 
		& \multirow{3}{*}{0.3}    & DIFFAI~\cite{mirman2018differentiable}      & \textgreater{24h}  & \textgreater{24$\times$} & N/A                                  \\
		&                         & Wong et al.~\cite{wong2018scaling}             & \textgreater{24h}   & \textgreater{24$\times$}   & N/A                                   \\
		&                         & \sys                                                   & {58m35s}                     & -                    & 96.6                                               \\ \hline
		\multirow{3}{*}{\begin{tabular}[c]{@{}c@{}}CIFAR\\ \_Small\end{tabular}}  & \multirow{3}{*}{0.0348} & DIFFAI~\cite{mirman2018differentiable}                                           & \textgreater{}24h               & \textgreater{16$\times$}                                                    & N/A                                              \\
		&                         & Wong et al.~\cite{wong2018scaling}                         & 3h35m                              & 3$\times$                                     & 61.3                                               \\
		&                         & \sys                                                          & {1h25m}       & -                                  & 69.4                                               \\ \hline
		\multirow{3}{*}{\begin{tabular}[c]{@{}c@{}}CIFAR\\ \_Large\end{tabular}}  & \multirow{3}{*}{0.0348} & DIFFAI~\cite{mirman2018differentiable}                         & \textgreater{24h}   & \textgreater{5$\times$}       & N/A                                   \\
		&                         & Wong et al.~\cite{wong2018scaling}                                           & 13h14m      & 3$\times$ & 64.9                             \\
		&                         & \sys (k=10)                                                                            & 4h28m             & -                                                       & 74.1                                               \\ \hline
		\multirow{3}{*}{\begin{tabular}[c]{@{}c@{}}CIFAR\\ \_Resnet\end{tabular}} & \multirow{3}{*}{0.0348} & DIFFAI~\cite{mirman2018differentiable}                      & \textgreater{24h}    & \textgreater{5$\times$}        & N/A                                               \\
		&                         & Wong et al.~\cite{wong2018scaling}          & 15h53m      & 3$\times$         & 60.2                                        \\
		&                         & \sys                                                             & 4h36m                                  & -                                  & {71.3}   \\\hline                 
	\end{tabular}
	\caption{\bf \small The shortest training time needed by different verifiable robust training schemes to achieve the same VRA as what \sys achieved in Table~\ref{tab:main2}, along with the corresponding test accuracy.}
	\label{tab:main3}
	\vspace{-5pt}
\end{table}

\subsection{\sys vs. state-of-the-art training schemes}
\label{sec:eval_train_schemes}
For each network architecture in Table~\ref{tab:networks}, we compared the performance of \sys
with existing state-of-the-art schemes: two adversarially robust training schemes by Madry et al.~\cite{madry2017towards} and Sinha et al.~\cite{sinha2018certifying} and two verifiably robust training schemes by Wong et al.~\cite{wong2018scaling} and Mirman et al. (DIFFAI)~\cite{mirman2018differentiable}.


For the two adversarially robust training schemes, we used the same adversary configurations and training process from Madry et al.~\cite{madry2017towards}. 
For the two verifiably robust training schemes, we picked the experimental setup that achieved the best
results as described in the corresponding paper. Wong et al.'s method~\cite{wong2018scaling} is the state-of-the-art at supporting verifiable robustness,
i.e., VRA. We used 50 random projections for Wong et al.'s method~\cite{wong2018scaling}. 
Similarly, we use training with box domains and testing with a hybrid Zonotope domain with transformer method hSwitch for DIFFAI~\cite{mirman2018differentiable} since this achieved the best results. 


\label{subsubsec:main1}

\noindent\textbf{Comparison with Adversarially Robust Training.} In Table~\ref{tab:main1}, we have shown the training results of different techniques for the five tested networks.
The best training time and VRA are highlighted for each network. Here, $k$ of \soa is set to be 1 to maximize the training process's efficiency.

When training the MNIST\_Large network with $\epsilon=0.3$ and CIFAR\_Resnet, Madry et al.'s method~\cite{madry2017towards}
does not converge. We cannot find a feasible choice of learning rate ($lr=10^{-7}, 3\times10^{-7},10^{-6},...,10$) that can train the network to have over 30\% test accuracy.

As shown by the results in Table~\ref{tab:main1}, \sys obtain the highest VRA for all networks. Madry et al.'s method~\cite{madry2017towards} is efficient but it cannot provide any verifiable robustness, i.e., has zero VRA. By contrast,  \sys can get VRA as high as 95.2\% while still being faster than Madry's method achieving same ACC and ERA. Such improvement in verifiable robustness is highly significant especially for security-sensitive applications. The training schemes of Sinha et al.~\cite{sinha2018certifying} can reach
higher ERA scores when $\epsilon=0.1$ compared to $\epsilon=0.3$ for MNIST networks. However, on CIFAR networks, it performed much worse in terms of ERA than \sys and Madry et al.'s method~\cite{madry2017towards}. This is consistent with the results reported in ~\cite{sinha2018certifying} as the distributional robustness works well for only very small $\epsilon$ values. Note that \sys is faster than the other two adversarially robust training schemes on all networks except CIFAR\_Large. Sinha et al.'s method~\cite{sinha2018certifying} is slightly faster than \sys on CIFAR\_Large, but we have significantly higher ACC, ERA, and VRA scores.

\label{subsubsec:main2}
\noindent\textbf{Comparison with Verifiably Robust Training.} In Table~\ref{tab:main2}, we compared the performance of \sys against verifiably robust training methods within 12-hours training time. Despite high VRA, currently the best verifiably robust trained networks provided by Wong et al.~\cite{wong2018scaling} still suffer from scalability issues and low test accuracy. Table~\ref{tab:main2} clearly demonstrates that \soa and \dmt are two very powerful methods that allow us to scale to large networks and balance the VRA and test accuracy.
Note that the results for DIFFAI on MNIST dataset are lower than the numbers reported in~\cite{mirman2018differentiable} mainly because their input normalization have a different scale (e.g., their $\epsilon=0.3$ only maps to $L_\infty$ norm of $25$ out of $255$ before normalization while our $\epsilon=0.3$ maps to $L_\infty=76$). Though DIFFAI with box domains are efficient to train, our experiments show that they tend to introduce large overestimation errors. Therefore, VRA of DIFFAI is low for most of the networks. Moreover, DIFFAI does not converge on MNIST\_Large and CIFAR\_Large networks.

While running \soa on small networks (MNIST\_Small, CIFAR\_Small), we set $k$ to be 20 to compute the verifiable robust loss without increasing the training cost too much. For the larger networks like CIFAR\_Resnet and CIFAR\_Large, we pick $k=10$. Note that we can make $k$ larger to further improve the VRA by spending more training time. For MNIST\_Large, we find that $k=1$ is enough to outperform the existing state-of-the-art.

For all networks, within 12-hour training time,\sys achieved higher ACC and VRA than DIFFAI and Wong et al.'s method . In particular, we have achieve 4.2\% higher VRA and 9.2\% higher ACC than the best reported number of Wong et al.~\cite{wong2018scaling} for the MNIST\_Large network ($\epsilon=0.3$).

\label{subsubsec:main3}
\noindent\textbf{Time to Achieve Target VRA.} We test whether other verifiably robust training methods can reach
the same VRA as \sys if we allow more training time. We set the training time threshold to be 24 hours and set the target VRA for all verifiably robust training methods to be those achieved by \sys in Table~\ref{tab:main2}.
The results are shown in Table~\ref{tab:main3}. 
\sys, on average,  is $15\times$ faster than Wong et al.'s method~\cite{wong2018scaling} and at least $20\times$ faster than DIFFAI for reaching the same target VRA.

\noindent\textbf{Results for ImageNet-200.} All existing verifiably robust training methods have so far been only evaluated on the MNIST and CIFAR dataset. Since the training cost of verifiably robust networks increase drastically for large datasets like ImageNet~\cite{imagenet_cvpr09,ILSVRC15}) and thus training networks with significant verifiable robustness is very hard. We evaluate \sys with ImageNet\_Resnet trained on the Imagenet-200~\cite{tiny-imagenet} dataset and show that \sys can scale to such large networks. We use robustness regions bounded by $L_\infty\leq 0.007$\footnote{the same robustness region is used in~\cite{mirman2018differentiable} for the CIFAR dataset.}. We set k=5 and train the networks for 12 hours. As shown in Table \ref{tab:imagenet}, \sys outperforms the state-of-the-art verifiable robust training methods by up to 11.8\% test accuracy and 14.4\% verified robust accuracy. 

\begin{table}[!hbt]
	\centering
	\small
	\begin{tabular}{cccc}
		\hline
		Method       & \multicolumn{1}{c}{\begin{tabular}[c]{@{}c@{}}Batch Time (s)\end{tabular}}     & ACC (\%) & VRA (\%)\\ \hline
		Regular training  & 0.022  & 36.2 & 0 \\ \hline
		Wong et al.~\cite{wong2018scaling} & 3.392     & 14.4        & 5.1       \\ \hline 
		\sys     & 0.396     & 26.2          & 19.4        \\ \hline
	\end{tabular}
	\caption{\bf \small The test accuracy (ACC) and verified robust accuracy (VRA) of ImageNet\_Resnet network trained with 12 hours using the ImageNet-200 dataset with $\epsilon\leq 0.007$.}
	\label{tab:imagenet}
\end{table}

\subsection{Robustness for different $L_\infty$ bounds}

\begin{figure}[!hbt]
	\centering
	\begin{subfigure}[t]{0.48\columnwidth}
		\includegraphics[width=\columnwidth]{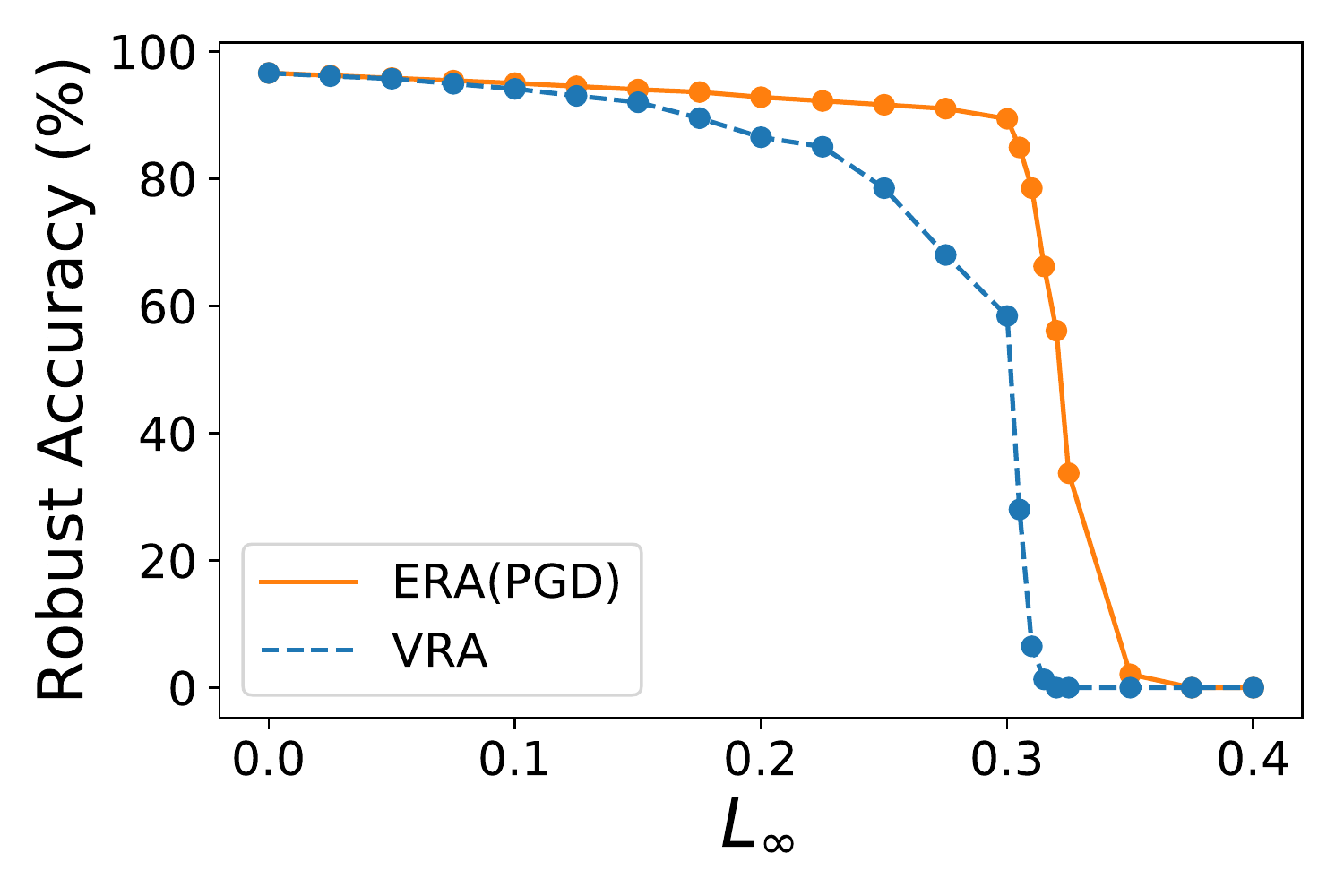}
		\caption{\bf \small MNIST\_Large}
		\label{subfig:minf2}
	\end{subfigure}
	\begin{subfigure}[t]{0.48\columnwidth}
		\includegraphics[width=\columnwidth]{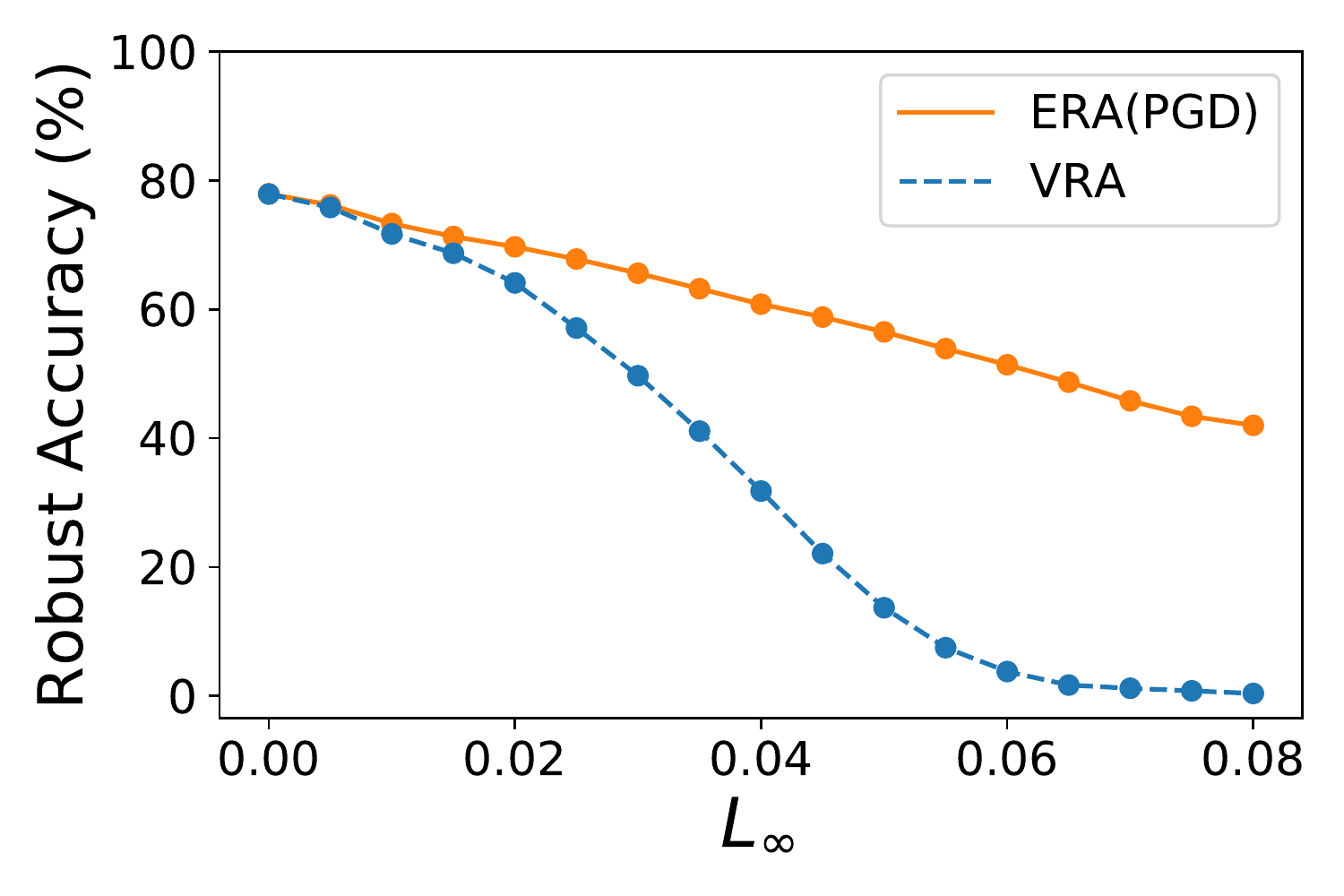}
		\caption{\bf \small CIFAR\_Large}
	\end{subfigure}
	\caption{\bf \small Changes of ERA and VRA in different $\epsilon$ regions of MNIST\_Large and CIFAR\_Large networks trained with \sys.}
	\label{fig:cinf}
	\vspace{-10pt}
\end{figure}

We examine the VRA and ERA metrics with different robustness regions in the CIFAR\_Large and MNIST\_Large networks trained with \sys. The CIFAR\_Large network and MNIST\_Large are trained with $L_\infty\leq 0.0348$ and $L_\infty\leq 0.3$, respectively. Figure \ref{fig:cinf} shows that these two networks have high VRA and ERA scores within large $L_\infty$ bounds. Overall, the ERA values are higher than VRA. For MNIST\_Large, the VRA score is still over 80\% when $\epsilon$ is $0.25$. Similarly for CIFAR\_Large, the VRA score is higher than 60\% when $\epsilon$ is $0.025$. Note that the verifiable robustness of MNIST\_Large decreases faster than CIFAR\_Large's, since the training range $L_\infty\leq 0.3$ is relative large that covers almost all visually undetectable perturbations.

\subsection{\sys vs. state-of-the-art attacks}

We show in Table \ref{tab:network_attack} that \sys can effectively defend against the state-of-the-art attacks including PGD, CW and interval attacks. For the same configurations of attacks, the interval attacks can find up to $1.4\times$ as many violations as PGD/CW attacks on adversarially robust trained networks as shown in Table \ref{tab:ia_mnist}. However, such advantage in success rate decreases to be less than 1\% on the networks trained with \sys. It indicates that, compared to Madry et al.'s method~\cite{madry2017towards}, \sys increases the verifiable robustness of networks to make them more robust and reliable under unknown attacks.

\begin{table}[!hbt]
	\small
	\centering
	\begin{tabular}{c@{\hskip7pt}c@{\hskip7pt}c@{\hskip7pt}c@{\hskip7pt}c@{\hskip7pt}c@{\hskip7pt}c}
		\hline
		Network & \begin{tabular}[c]{@{}c@{}}ACC\\ (\%)\end{tabular} & \begin{tabular}[c]{@{}c@{}}ERA\\(PGD) (\%)\end{tabular} & \begin{tabular}[c]{@{}c@{}}ERA\\(CW)\end{tabular}(\%)  & \begin{tabular}[c]{@{}c@{}}ERA\\(IA)$^*$\end{tabular}(\%)  & \begin{tabular}[c]{@{}c@{}}VRA\\ (\%)\end{tabular} \\ \hline
		MNIST\_Small  & 95.4               & 87.3          &  87.2            &   86.7             & 52.0       \\ \hline
		MNIST\_Large  & 96.6               & 89.4          &  90.0            &   88.6             & 58.4     \\ \hline
		CIFAR\_Small  & 71.1               & 54.6          &   53.9           &   53.1             & 37.6     \\ \hline
		CIFAR\_Large  & 77.9               & 63.5          &   62.8           &   62.2             & 41.6     \\ \hline
		CIFAR\_Resnet & 77.5               & 61.0          &   60.4           &   59.6             & 35.2       \\ \hline
		\multicolumn{6}{l}{\footnotesize * Estimated robust accuracy under interval attacks.}
	\end{tabular}
	\caption{\bf \small ERA of networks trained with \sys under different attacks ($\epsilon$ is 0.3 for MNIST and 0.0348 for CIFAR).}
	\label{tab:network_attack}
	\vspace{-10pt}
\end{table}

\subsection{\sys against $L_0$ and $L_2$ bounded attacks}
\begin{figure}
	\centering
	\begin{subfigure}[t]{0.49\columnwidth}
		\includegraphics[width=\columnwidth]{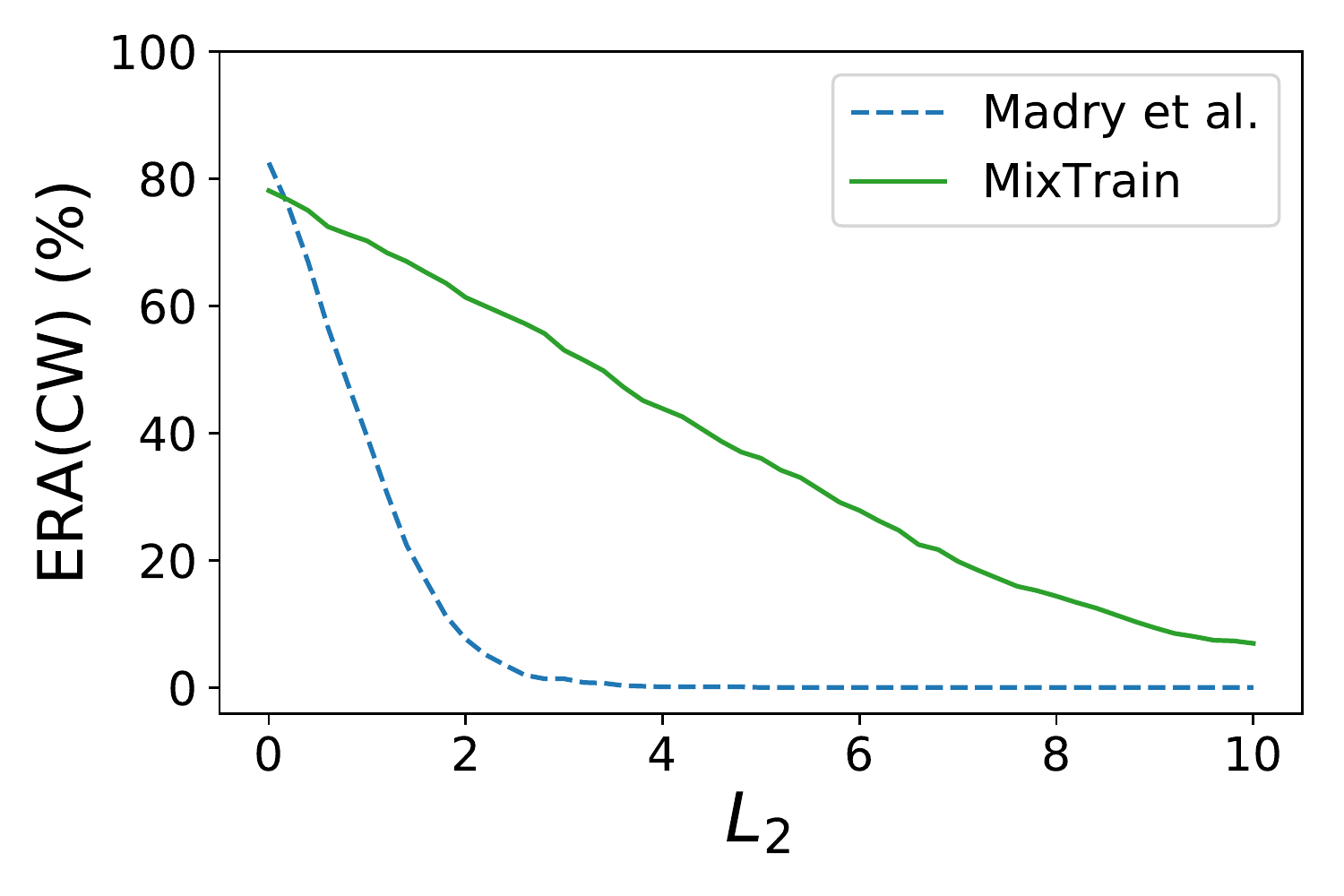}
		\caption{\bf \small $L_2$ distance}
		\label{subfig:l2}
	\end{subfigure}
	\begin{subfigure}[t]{0.49\columnwidth}
		\includegraphics[width=\columnwidth]{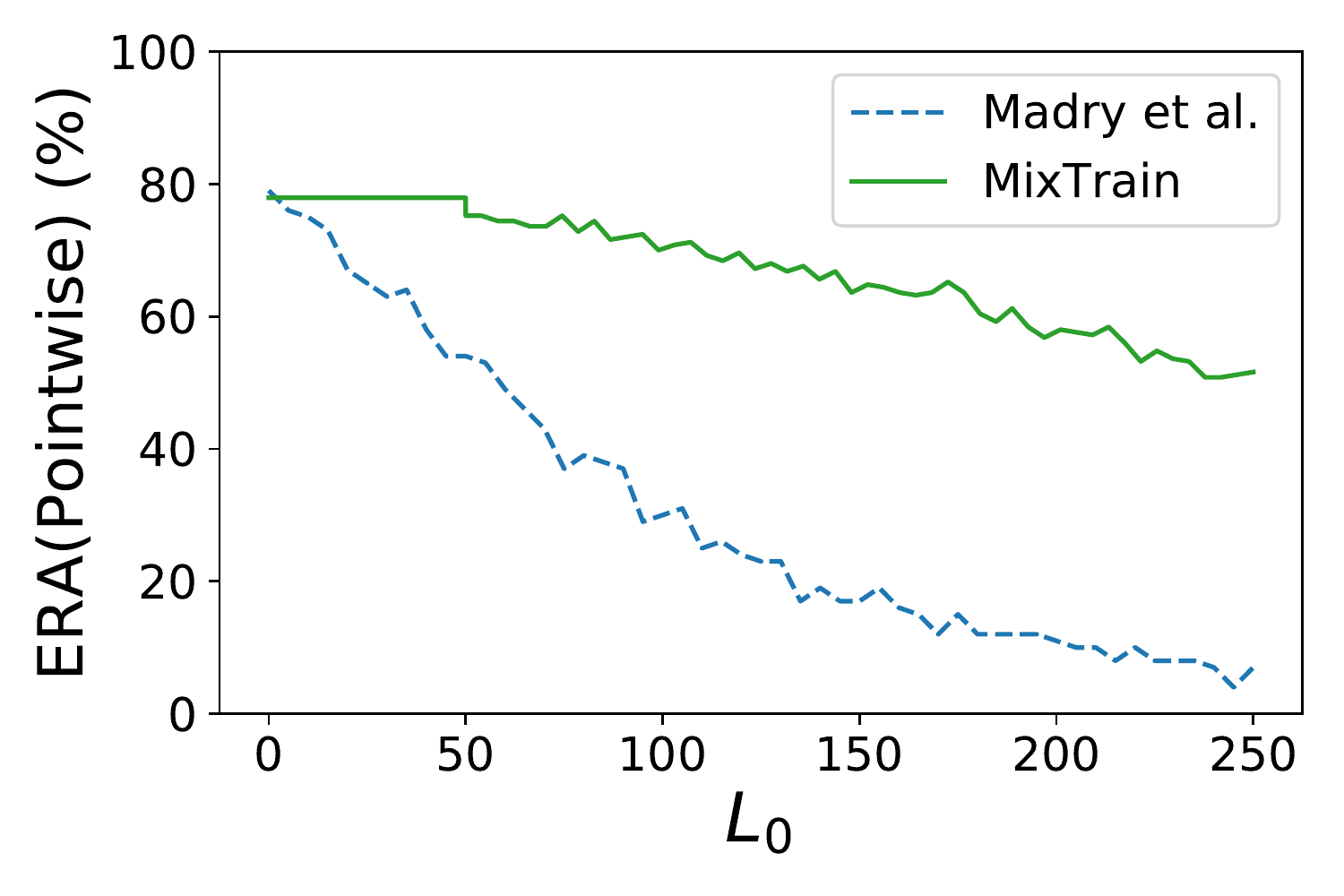}
		\caption{\bf \small $L_0$ distance}
		\label{subfig:l0}
	\end{subfigure}
	\caption{\bf \small ERA of CIFAR\_Large networks trained with \sys vs Madry et al.~\cite{madry2017towards}, under $L_2$-bounded CW attacks and $L_0$-bounded pointwise attacks. As additional benefits, the networks trained with \sys are much more robust against $L_0$ and $L_2$ adversaries.}
	\label{fig:cl2l0}
	\vspace{-10pt}
\end{figure}

Using $L_\infty$-bound for verifiably robust training, \sys can also make networks robust against $L_0$ and $L_2$-bounded attacks.
Schott et al.~\cite{lucas2018towards} have shown that Madry et al.'s method~\cite{madry2017towards} is not robust against other types of attacks like $L_0$ and $L_2$-bounded attacks. In Figure \ref{fig:cl2l0}, we measure the ERA under state-of-the-art $L_2$-bounded CW attacks and $L_0$-bounded decision-based pointwise attacks~\cite{lucas2018towards,rauber2017foolbox} on CIFAR\_Large networks trained with \sys. We compare the differences of the ERA obtained through \sys against the networks trained from Madry et al.'s method~\cite{madry2017towards}. For $L_2$-bounded CW attacks, as shown in Figure \ref{subfig:l2}, the ERA of adversarially robust trained network quickly decreases to $0\%$ when $L_2= 2$, whereas \sys can retain over $60\%$ ERA. For $L_0$-bounded pointwise attacks, as shown in Figure \ref{subfig:l0}, ERA of the network trained with \sys stays over $50\%$ at $\epsilon=250$ while Madry et al.'s number drops to $0\%$. The results demonstrate that the networks trained with \sys are much more robust against $L_0$ and $L_2$ adversaries as side benefits of \sys under $L_\infty$-bounded sound over-approximations.

\subsection{Different values of sampling rate $k$}

We evaluate different choices of $k$ in \soa over the resulting accuracy metrics and training time in Table~\ref{tab:k}. Larger $k$ values can increase ERA and VRA scores at the cost of more training time. In particular, we train the CIFAR\_Small networks with \sys on one GPU. When $k=1$, the network achieves 71.1\% test accuracy and decent VRA (37.6\%), with very efficient training time. Therefore, $k=1$ is a good choice to save training time/memory. On the other hand, if more training time is available, one can use larger sampling rate $k$ to obtain higher VRA. For instance, with $k=20$, \sys can train a robust network with a high VRA ($45.4\%$ vs $47.5\%$) and a significantly higher test accuracy ($70.7\%$ vs $61.1\%$) compared to the state-of-the-art method by Wong et al.~\cite{wong2018scaling}. Also, it only takes \sys 2 hours and 42 minutes to train whereas existing verifiable robust training techniques need over 11 hours to reach the same VRA.

\begin{table}[!hbt]
	\centering
	\small
	\begin{tabular}{c@{\hskip8pt}rr@{\hskip10pt}c@{\hskip10pt}c@{\hskip8pt}c}
		\hline
		& \begin{tabular}[c]{@{}c@{}}Batch\\Time\end{tabular} (s) & \begin{tabular}[r]{@{}r@{}}Training\\Time\end{tabular} & ACC (\%) & \begin{tabular}[c]{@{}c@{}}ERA\end{tabular} (\%) & VRA (\%) \\ \hline
		k=1  & 0.015          & 14m52s        & 71.1               & 54.6          & 37.6     \\ \hline
		k=5  & 0.042          & 45m23s        & 70.7               & 54.9          & 39.9     \\ \hline
		k=10 & 0.078          & 1h18m            & 69.7               & 55.2          & 43.2     \\ \hline
		k=20 & 0.162          & 2h42m        & 70.7               &  56.7         & 45.4    \\ \hline
	\end{tabular}
	\caption{\bf \small Under different values of sampling rate $k$ for \soa, the comparisons of corresponding training time, ACC, ERA, and VRA on CIFAR\_Small networks.}
	\label{tab:k}
\end{table}

\subsection{Different values of $\alpha$}
\label{subsec:eval_alpha}

\begin{figure}
	\centering
	\includegraphics[width=0.7\columnwidth]{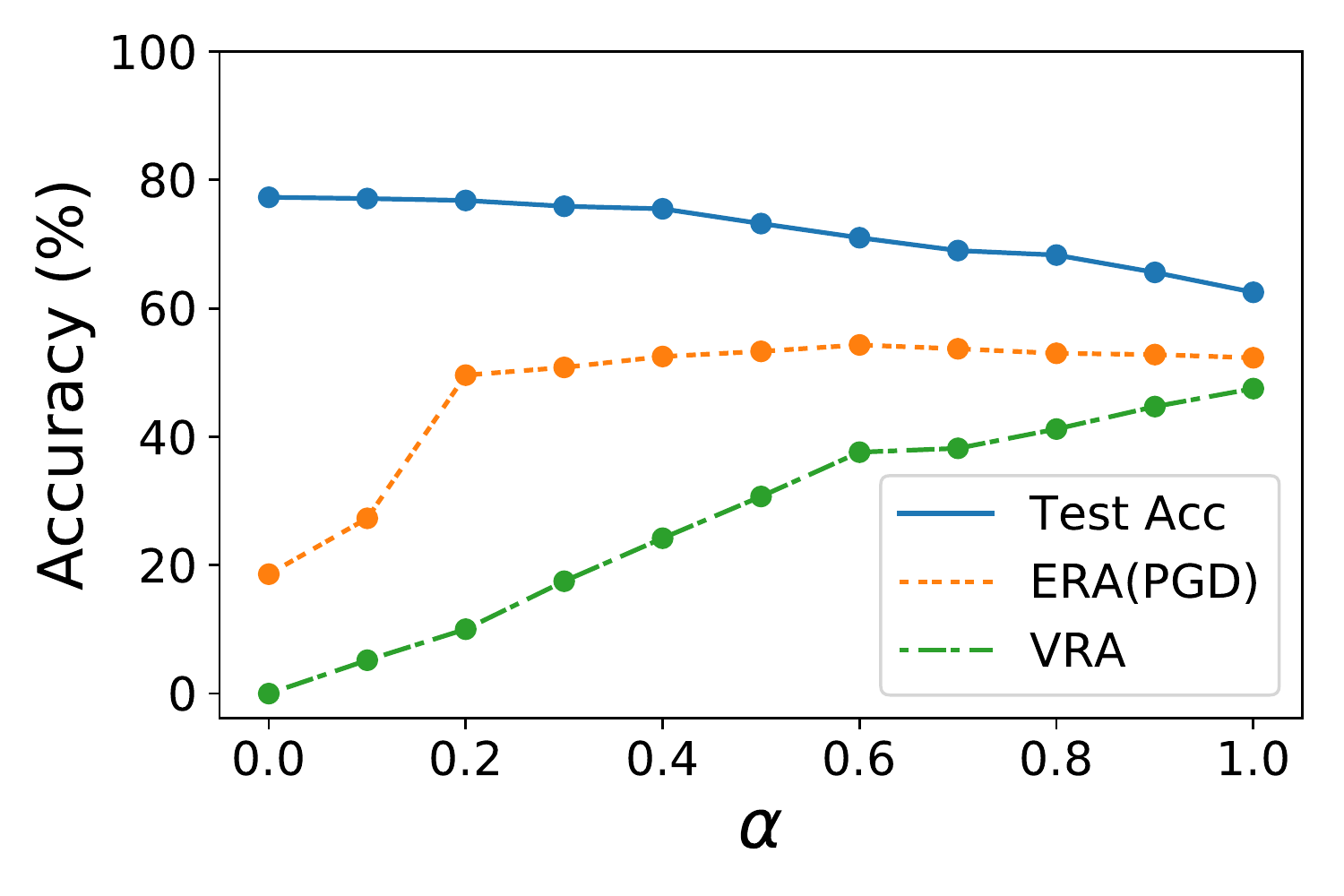}
	\caption{\bf \small ACC, ERA, and VRA of CIFAR\_Small networks trained with \sys under different fixed $\alpha$ values in \dmt.}
	\label{fig:alpha}
	\vspace{-15pt}
\end{figure}

\noindent\textbf{Influence of different $\alpha$ values.} We measure how ACC, ERA, and VRA change with different fixed $\alpha$ values used for training the CIFAR\_Small network. Specifically, we use different constant values $\alpha=0.0,0.1,...,1.0$ to evaluate the corresponding metrics of each network trained with $k=1$. Figure~\ref{fig:alpha} shows that as $\alpha$ increases, both ERA and VRA increase but test accuracy decreases. If $\alpha=0$, the training process is the same as regular training, which has the highest ACC but $0\%$ VRA. By contrast, if $\alpha=1$, \sys only relies on \soa and loses the ACC improvement provided by \dmt. Overall, by adjusting $\alpha$, \sys can be tuned to balance the verifiable robustness and test accuracy as desired.

\noindent\textbf{Dynamic update of $\alpha$.} Instead of using a fixed $\alpha$ value in training, dynamically changing $\alpha$ can achieve higher ERA and VRA. In Table~\ref{tab:fix}, we compare the results achieved by dynamic $\alpha$ with those by a fixed $\alpha$. First, we train with a fixed $\alpha=0.5$ that balances between ACC and VRA of the network. Next, we adaptively update $\alpha$ as described in Section~\ref{subsubsec:alpha} and compare the resulting ACC, ERA, and VRA metrics.  As shown in Table \ref{tab:fix}, the network trained with dynamic $\alpha$ outperforms the network trained under fixed $\alpha$ mechanisms by $6.7\%$ VRA while achieving similar ACC.

\begin{table}[!hbt]
	\small
	\centering
	\begin{tabular}{|c|c|c|c|}
		\hline
		CIFAR\_Small   & ACC (\%) & ERA (\%) & VRA (\%) \\ \hline
		Fixed $\alpha=0.5$      & 71.4               & 53.2          & 30.9     \\ \hline
		Adaptive $\alpha$ & 71.7               & 54.6          & 37.6     \\ \hline
	\end{tabular}
	\caption{\bf \small Performance of CIFAR\_Small networks trained with dynamically adapted $\alpha$ vs fixed $\alpha$.}
	\label{tab:fix}
\end{table}

\section{Related Work}
\label{sec:related}

Recent works have shown that the state-of-the art neural networks can be easily fooled with adversarially crafted human-imperceptible perturbations on valid inputs~\cite{carlini2017towards,szegedy2013intriguing,goodfellow2014explaining,moosavi2016deepfool}. The proposed defenses can be categorized into two classes\textemdash best-effort defenses and principled defenses satisfying some security properties.

\noindent\textbf{Best-effort defenses.}
Defenses of this class usually rely on heuristics without supporting any clear security guarantees about the robustness of the underlying network to adversarial inputs~\cite{gu2014towards,papernot2015distillation,cisse2017parseval,papernot2017extending,papernot2018deep,athalye2018obfuscated,buckman2018thermometer,song2017pixeldefend,xie2017mitigating,zantedeschi2017efficient}. Therefore, these defenses  have been followed by a sequence of stronger attacks breaking them in quick succession~\cite{papernot2016cleverhans,carlini2017towards,elsayed2018adversarial,carlini2017adversarial,moosavi2016deepfool,biggio2013evasion,ma2018characterizing,papernot2017practical,guo2017countering,he2017adversarial,athalye2017synthesizing,carlini2018efficient,pei2017deepxplore,tian2018deeptest}. We refer the interested readers to the survey by Athalye et al.~\cite{athalye2018obfuscated} for more details. 


\noindent\textbf{Principled defenses.}
Adversarially robust training and verifiably robust training are the two main classes of principled defenses.
Adversarially robust training uses various different existing attacks~\cite{goodfellow2014explaining,kurakin2016adversarial, tramer2017ensemble,ducoffe2018adversarial, ma2018characterizing, madry2017towards} to make the trained network robust. Overall, they are robust against existing attacks~\cite{madry2017towards}, but can still be defeated by smarter attackers as shown in this paper.

Verifiably robust training can provide sound estimates of the robustness of the trained network. They use sound over-approximation of violations to guide the training process. The underlying techniques for soundly measuring robustness include basic linear programming~\cite{lomuscio2017approach}, SMT solvers~\cite{katz2017reluplex,huang2017safety,ehlers2017formal}, MILP solvers~\cite{tjeng2017evaluating,fischetti2017deep,dutta2018output}, semidefinite relaxations~\cite{raghunathan2018certified}, symbolic intervals~\cite{reluval2018,shiqi2018efficient}, convex polytopes~\cite{wong2018provable}, abstract domains~\cite{gehrai}, Lagrangian relaxations~\cite{dvijotham2018dual} and Lipschitz continuity~\cite{weng2018evaluating,weng2018towards}. To increase the efficiency of the training process, verifiably robust training often has to make some sacrifice in the soundness of the robustness estimates~\cite{dvijotham2018training,mirman2018differentiable}. Currently, the state of the art, leveraging random projections to save computations~\cite{wong2018scaling}, is still hundreds of times slower than regular training. \sys speeds up the verifiably robust training by on average 15 times with up to 13.2\% test accuracy improvement.

\section{Conclusion}
\label{discussion}

In this paper, we designed, implemented, and evaluated \soa and \dmt as part of \sys for making verifiably robust training of neural networks an order of magnitude faster. Our extensive experimental results demonstrated that \sys outperforms the state-of-the-art verifiably robust training methods achieving 13.2\% improvement in test accuracy and up to 6.5\% higher verified robust accuracy while taking $36\times$ less training time and $50\times$ less memory. \sys even takes up to 2x less training time than the state-of-the-art adversarially robust training methods that cannot  provide any verified robustness accuracy. We further presented a new first-order attack, the interval attack, that shows the fundamental limitations of adversarially robust training~\cite{madry2017towards}. By contrast, \sys provides strong security any L-p-norm bounded attacks including interval attacks.

\bibliographystyle{IEEEtran}
\bibliography{ref}

\appendix
\vspace{-6pt}
\subsection{Memory Usage}
\label{subsec:mem}

\sys is able to significantly cut down memory usage during robust training compared to state-of-the-art verifiable robust training~\cite{wong2018scaling}. This allows \sys to run on limited GPU resources. Here, in Table \ref{tab:mem}, we summarize the GPU memory required for regular training, \sys and Wong et al.~\cite{wong2018scaling} for each network. $\epsilon$ is set to 0.3 for MNIST dataset and 0.0348 for CIFAR dataset. For Wong et al.'s mehtod~\cite{wong2018scaling}, even after applying 50 random projection proposed in~\cite{wong2018scaling} to cut down memory usage, it still needs 4 GeForce GTX 1080 Ti GPUs to train large networks with batch size 50. In contrast, \sys requires only 1 GPU to train these networks. Overall, \sys, on average, uses $10\times$ less GPU memory than Wong et al.'s method. Note that, as shown in Table \ref{tab:main2} and \ref{tab:main3}, \sys further provides similar or even better accuracy and robustness in less training time compared to Wong et al.'s method~\cite{wong2018scaling}.

\begin{table}[!hbt]
	\centering
	\small
	\begin{tabular}{cccc}
		\hline
		\multirow{2}{*}{Network} & \multicolumn{3}{c}{Memory Usage (MB)}\\ 
		       & Baseline   & \sys &  Wong et al.~\cite{wong2018scaling} \\ \hline
		MNIST\_Small     & 533      & 837           & 5,474 \\ \hline
		MNIST\_Large     & 629     & 6,383          &   32,626        \\ \hline
		CIFAR\_Small     & 553     & 869        & 17,960       \\ \hline 
		CIFAR\_Large     & 621     & 5,335          &  39,840        \\ \hline
		CIFAR\_Resnet    & 589     & 3,475          &  3,7784   \\ \hline
	\end{tabular}
	\caption{\bf \small The GPU memory usage required to train different networks by using \sys compared to regular training (Baseline) and state-of-the-art verifiable robust training method~\cite{wong2018scaling}. $\epsilon$ is 0.3 for MNIST dataset and 0.0348 for CIFAR dataset.}
	\label{tab:mem}
\end{table}

\vspace{-15pt}
\subsection{Network Architectures}
\label{sec:ans}

\newcommand{\relu}{\text{ ReLU }}

Below we describe the architectures of the 6 different networks we used for our evaluation. We use $Conv_sC\times W\times H$ to denote a convolutional layer that has $C$ channels with kernel size $(W, H)$ and stride $s$.  $FC\text{ }n$ denotes a fully connected layer that contains $n$ hidden nodes and $Res\text{ }C\times W\times H$ denotes the residual modules defined in~\cite{zagoruyko2016wide} that has $C$ output channels with kernel size $(W,H)$. Each residual module contains two convolutional layers each with ReLU and one skip connection from input. 

\noindent\textbf{MNIST\_FC1.} Small fully connected network for MNIST:
\vspace{-5pt}
\begin{equation*}
\begin{split}
x &\rightarrow FC\text{ }512\rightarrow \relu \rightarrow FC\text{ }512\rightarrow \relu \rightarrow f_\theta(x)
\end{split}
\end{equation*}

\vspace{-3pt}\noindent\textbf{MNIST\_FC2.} Large fully connected network for MNIST:
\vspace{-5pt}
\begin{equation*}
\begin{split}
x &\rightarrow FC\text{ }2048\rightarrow \relu \rightarrow FC\text{ }2048\rightarrow \relu \\&\rightarrow FC\text{ }2048\rightarrow \relu \rightarrow FC\text{ }2048\rightarrow \relu\\&\rightarrow FC\text{ }2048\rightarrow \relu \rightarrow f_\theta(x)
\end{split}
\end{equation*}

\vspace{-3pt}\noindent\textbf{MNIST\_Small \& CIFAR\_Small.} Small convolutional network for MNIST and CIFAR used in~\cite{wong2018provable}:
\vspace{-5pt}
\begin{equation*}
\begin{split}
x &\rightarrow Conv_2 16\times4\times4\rightarrow \relu \rightarrow Conv_2 32\times4\times4 \\& \rightarrow \relu  \rightarrow FC\text{ }100\rightarrow \relu \rightarrow f_\theta(x)
\end{split}
\end{equation*}
\vspace{-3pt}

\vspace{-3pt}\noindent\textbf{MNIST\_Large \& CIFAR\_Large.} Large convolutional networks that are the extensions of small convolutional ones:
\vspace{-5pt}
\begin{equation*}
\begin{split}
x &\rightarrow Conv_1 32\times3\times3\rightarrow \relu \rightarrow Conv_2 32\times4\times4  \\ &\rightarrow \relu \rightarrow Conv_1 64\times3\times3\rightarrow \relu\\ &\rightarrow Conv_2 64\times4\times4\rightarrow \relu  \rightarrow FC\text{ }512 \\ &\rightarrow \relu \rightarrow FC\text{ }512\rightarrow \relu \rightarrow f_\theta(x)
\end{split}
\end{equation*}

\vspace{-3pt}\noindent\textbf{CIFAR\_Resnet.} The residual networks use the same in~\cite{zagoruyko2016wide}:
\vspace{-5pt}
\begin{equation*}
\begin{split}
x &\rightarrow Conv_1 16\times3\times3 \rightarrow \relu\rightarrow Res\text{ }16\times3\times3 \\ &\rightarrow Res\text{ }32\times3\times3 \rightarrow Res\text{ }32\times3\times3 \\ &\rightarrow Res\text{ } 64\times3\times3 \rightarrow Res\text{ } 64\times3\times3\\ & \rightarrow FC\text{ }1000\rightarrow \relu \rightarrow f_\theta(x)
\end{split}
\end{equation*}

\vspace{-3pt}\noindent\textbf{ImageNet\_Resnet.} The residual networks use the same structure as in~\cite{zagoruyko2016wide} for Imagenet-200 dataset~\cite{tiny-imagenet}. To improve training efficiency, we crop the original shape of image 3$\times$64$\times$64 to 3$\times$32$\times$32. Such data augmentation help to speed up the the training procedure without sacrificing too much accuracy.
\vspace{-5pt}
\begin{equation*}
\begin{split}
x &\rightarrow Conv_1 16\times3\times3 \rightarrow \relu\rightarrow Res\text{ }16\times3\times3 \\ &\rightarrow Res\text{ }32\times3\times3 \rightarrow Res\text{ }32\times3\times3 \\ &\rightarrow Res\text{ } 64\times3\times3 \rightarrow Res\text{ } 64\times3\times3\\ & \rightarrow FC\text{ }1000\rightarrow \relu \rightarrow f_\theta(x)
\end{split}
\end{equation*}

\end{document}